\pdfoutput=1
\documentclass[11pt]{article}

\usepackage[table]{xcolor}
\usepackage[preprint]{acl}

\usepackage{fancyhdr}
\fancypagestyle{firstpage}{
  \fancyhf{}
    \fancyhead[L]{\normalsize Accepted in EMNLP 2025}
     \fancyfoot[C]{\thepage}
}

\usepackage{times}
\usepackage{latexsym}

\usepackage[T1]{fontenc}
\usepackage[utf8]{inputenc}

\usepackage{microtype}

\usepackage{inconsolata}

\usepackage{graphicx}
\usepackage{booktabs}
\usepackage{amssymb}
\usepackage{amsmath}
\usepackage{multirow}
\usepackage{array}
\usepackage{tabularx}
\usepackage{arydshln} % For dashed lines
\usepackage{caption}
\usepackage{hyperref}

\title{RCI: A Score for Evaluating Global and Local Reasoning in Multimodal Benchmarks}

\author{
\textbf{Amit Agarwal\textsuperscript{}},
\textbf{Hitesh Laxmichand Patel\textsuperscript{}},
\textbf{Srikant Panda\textsuperscript{}},
\textbf{Hansa Meghwani\textsuperscript{}}, \\
\textbf{Jyotika Singh\textsuperscript{}},
\textbf{Karan Dua\textsuperscript{}}, 
\textbf{Paul Li\textsuperscript{}}, 
\textbf{Tao Sheng\textsuperscript{}},
\textbf{Sujith Ravi\textsuperscript{}},
\textbf{Dan Roth\textsuperscript{}}
\\ 
\\
 \textsuperscript{}Oracle AI
\\
 \small{
   \textbf{Correspondence:} \href{mailto:email@domain}{amit.h.agarwal@oracle.com}
 }
}

\begin{document}
\maketitle

\thispagestyle{firstpage}
\pagestyle{firstpage}

\begin{abstract}
Multimodal Large Language Models (MLLMs) have achieved impressive results on vision-language benchmarks, yet it remains unclear whether these benchmarks assess genuine global reasoning or allow success via localized visual cues. Existing evaluation methods do not explicitly measure this distinction, hindering effective dataset curation and real-world focused model development.

We introduce \textbf{Region Comprehension Index (RCI)}, the first model-based score to directly quantify a dataset’s reliance on global versus local visual information. RCI systematically compares reference-model performance on image patches versus full images, revealing if tasks require holistic image understanding or can be solved with partial or localized visual cues.

When applying RCI to 13 widely used multimodal benchmarks, we observed that most of them favor localized reasoning and exhibit significant spatial biases, indicating potential risks in real-world applications. RCI equips researchers \& practitioners with an actionable tool for diagnosing \& mitigating these biases, enabling the construction of datasets and benchmarks to foster the development of robust, enterprise-ready multimodal systems.

\end{abstract}

% ACL 
% \begin{abstract}
% Multimodal Large Language Models (MLLMs) have shown remarkable performance across various vision-language benchmarks. However, it remains unclear whether these benchmarks primarily assess model's global reasoning capabilities on a whole image or rely on localized visual cues. Existing evaluation frameworks do not explicitly measure this distinction, making it difficult to choose appropriate data sets or design targeted model training strategies. To address this, we introduce the Region Comprehension Index (RCI), a novel score designed to quantify datasets \& benchmarks reliance on global versus local reasoning. By systematically comparing model performance on individual image patches versus entire images, RCI identifies dataset-induced biases toward localized reasoning. 

% Evaluating 13 popular multimodal benchmarks, we find that six predominantly favor localized reasoning, only three strongly require global context, while the rest exhibit a mix of local and global reasoning requirement. RCI provides dataset curators and benchmark designers with a practical tool to explicitly assess reasoning dependencies, enabling the creation of datasets that foster robust and generalizable multimodal models suited for real-world applications.
% \end{abstract}

\section{Introduction}

MLLMs have driven dramatic progress in vision-language tasks such as visual question answering \cite{pattnayak2024survey,pattnayak2025hybrid}, image captioning, and data generation \cite{agarwal2024synthetic,agarwal2024enhancing,patel2024llm,patel2025sweeval}, enabled by advances in model architectures and the availability of large-scale datasets and standardized benchmarks. Yet, as these models transition from academic labs to real-world applications, a critical question arises: \textit{Do current benchmarks truly reflect the reasoning demands of practical, deployment-oriented systems, or do they enable models to succeed via narrow, localized cues?}

Recent studies~\cite{paper3,paper4} reveal that many popular benchmarks allow high performance through exploitation of limited or localized visual context, without requiring genuine integration of global visual information across the image. This often results in models that appear robust on paper, but fail to generalize or perform reliably when deployed in real-world settings.

In industrial and mission-critical applications, such as autonomous driving, remote sensing, medical imaging, document intelligence, and large-scale content moderation, models must demonstrate robust \textit{global reasoning}: correlating information distributed across an entire image or scene. Conversely, some practical tasks (e.g., facial recognition, fine-grained inspection, anomaly detection) require only highly \textit{localized reasoning}: analyzing localized visual cues. A persistent challenge is that existing benchmarks rarely make these reasoning dependencies explicit, leading to costly misalignment between evaluation metrics and real-world system requirements.

This ambiguity stems from the design of current benchmarks, many of which can be solved by models leveraging local visual features or cues, creating an illusion of general visual reasoning. This misalignment risks wasted effort and, more importantly, unreliable deployed systems.

In this work, we propose the \textit{Region Comprehension Index} (RCI), a practical, model-based score for auditing and guiding the development of multimodal benchmarks and datasets. RCI systematically compares a reference-model performance on individual image patches versus full images, across different granularities, to quantify whether a dataset requires global (holistic) or local reasoning to succeed. In contrast to traditional evaluation metrics (e.g., FID~\cite{fid_score}, CLIPScore~\cite{hessel2021clipscore}, CIDEr~\cite{vedantam2015ciderconsensusbasedimagedescription}), which primarily focus on alignment or diversity, RCI provides an actionable signal for both researchers and industry practitioners to diagnose, compare, and curate benchmarks that better match real-world deployment needs. Our key contributions are:
\begin{itemize}
    \item Introducing RCI, the first score to explicitly quantify global vs.\ local reasoning requirements in multimodal benchmarks.
    \item Presenting a structured, patch-based evaluation framework to reveal and analyze spatial reasoning biases in vision-language datasets.
    \item Applying RCI to 13 widely-used benchmarks, providing %actionable
    insights for data \& system designers in both research \& industry contexts.
\end{itemize}

By enabling practitioners to audit and align benchmarks with actual application demands, RCI helps bridge the gap between academic evaluation and real-world deployment, supporting the development of robust, generalizable multimodal systems.

\section{Related Work}
\label{sec:related_work}

\subsection{Vision-Language Benchmarks}%[CAMERA READY VERSION] and Dataset Biases}
Vision-language benchmarks such as MS COCO~\cite{chen2015microsoft}, GQA~\cite{hudson2019gqanewdatasetrealworld}, TextVQA \& VizWiz have significantly advanced multimodal model development. However, recent works~\cite{geirhos2020shortcut, guan2024hallusionbenchadvanceddiagnosticsuite,paper3,paper4,whatsup2024} reveal critical limitations: many tasks can be effectively addressed by exploiting localized visual information, creating an \textit{illusion of progress}. For instance, models often leverage minimal contextual clues \& biased spatial distributions to achieve deceptively high benchmark scores~\cite{paper3,whatsup2024}. Benchmarks like SPEC~\cite{spec2024}, AMBER, BLINK, MVTamperBench \cite{agarwal-etal-2025-mvtamperbench}, \& What’s Up~\cite{whatsup2024} explicitly highlight these issues by isolating fine-grained spatial-temporal \& semantic reasoning tasks, uncovering significant model limitations. Such localized shortcuts undermine robustness, interpretability, \& generalization, %particularly 
impacting real-world applications like medical analysis \cite{pattnayak2025clinicalqa20multitask}, document analysis \cite{meghwani-etal-2025-hard,agarwal-etal-2025-fs}, \& autonomous navigation, where comprehensive visual reasoning is essential.
 % [CAMERA READY VERSION]However, these studies do not provide a unified approach for systematically quantifying global contextual reasoning capabilities across diverse tasks, emphasizing the need for a generalized evaluation framework.

\subsection{Spatial Reasoning and Dataset Quality Assessment}
Spatial reasoning remains a challenging yet essential capability for vision-language models~\cite{wu2024vspassessingdualchallenges,whatsup2024}. Recent research demonstrates widespread deficiencies in spatial relation comprehension, even among advanced models~\cite{whatsup2024}. For example, SPEC explicitly diagnoses model comprehension of spatial attributes, demonstrating near-random performance even for state-of-the-art MLLMs ~\cite{spec2024}. Similarly, \citet{zhaoarticle} highlight the importance of dataset quality, revealing substantial annotation issues that exacerbate spatial reasoning deficiencies. To address these shortcomings, some researchers have proposed visual prompting techniques, guiding model's attention explicitly through visual cues~\cite{yu2024attention}. %[CAMERA READY] Although these methods enhance task performance, they do not systematically quantify reasoning dependencies or evaluate spatial biases inherent in benchmarks.

\begin{figure*}[th!]
    \centering
    \includegraphics[width=\textwidth]{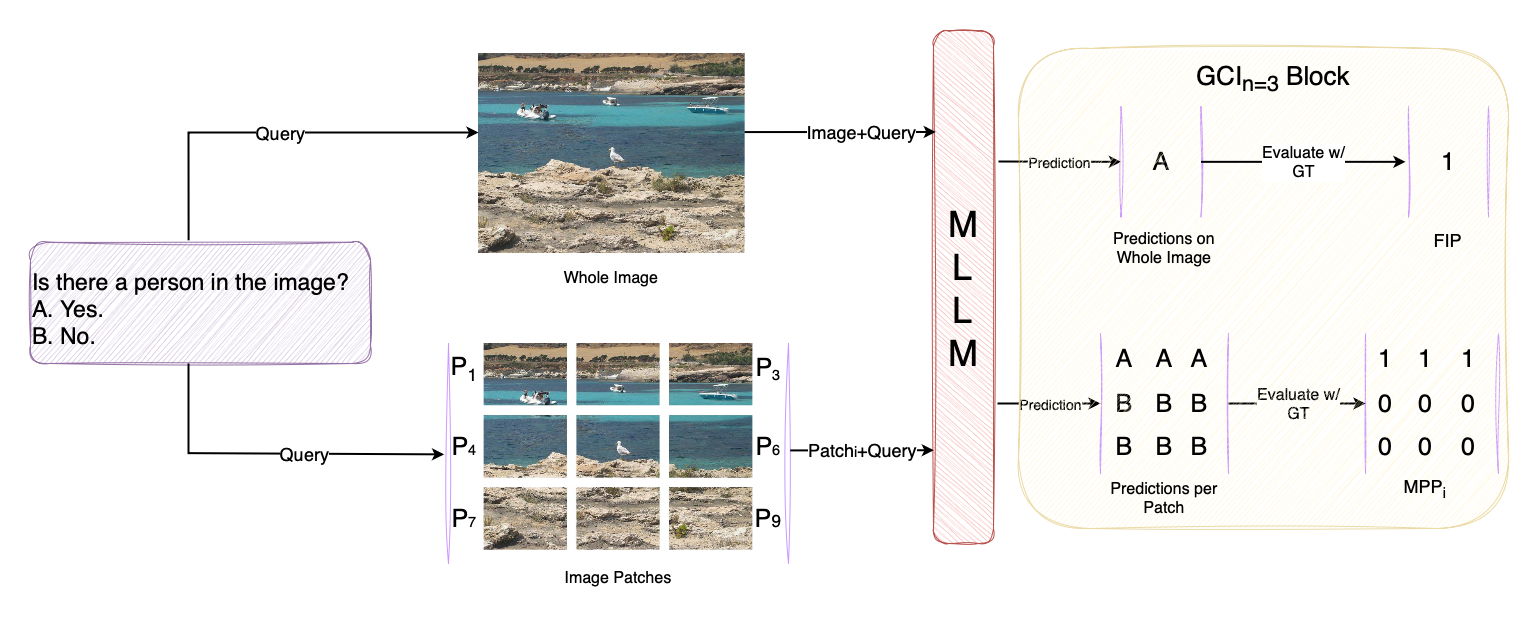}
    % \caption{Illustration of a sample image from the POPE benchmark, on how we compute the FIP (on the top) and MPP (on the bottom) for a single image. We then aggregate the predictions to compute the performance on the entire benchmark dataset.}
    \caption{Computation of Full Image Performance (FIP) and Maximum Patch Performance (MPP) on a sample from the POPE benchmark. FIP (top) evaluates model performance on the full image, while MPP (bottom) identifies the highest-performing patch. These are aggregated across the dataset to compute RCI.%, quantifying a benchmark’s reliance on global vs. localized reasoning.
}
    % Later we aggregate ( \(\sum_{i}\) ) them over the entire dataset for computation.}
    \label{fig:RCIn_framework}
    \vspace{-1.2em}
\end{figure*}

% \subsection{Evaluation Metrics for Vision-Language Tasks}
\subsection{Metrics for Vision-Language Tasks}
Traditional metrics like CIDEr, BLEU~\cite{papineni2002bleu}, and METEOR~\cite{banerjee2005meteor} primarily assess linguistic properties such as caption similarity and diversity, but fail to explicitly capture deeper reasoning capabilities or spatial dependencies. Metrics like CLIPScore and FID evaluate cross-modal alignment and image realism. % [CAMERA READY] but do not explicitly quantify spatial or reasoning biases inherent in datasets.

Recently, \citet{tao2024probingmultimodallargelanguage} probed multimodal models for global and local semantic representations, highlighting discrepancies in representation across model layers. Although insightful, this work does not quantify how datasets themselves structure or promote spatial reasoning explicitly.

% \subsection{Positioning Our Work}
\paragraph{Our Contribution}
Our research explicitly addresses these critical gaps by introducing RCI. Unlike previous approaches that focus on isolated evaluation dimensions or linguistic alignment, RCI systematically measures and reveals whether a dataset’s tasks fundamentally depend on integrating information across the entire image, or can be addressed using isolated regions. %RCI is a model-based dataset audit that complements standard accuracy-style metrics by exposing the type of visual information a dataset rewards (global vs. local). It is intended to guide benchmark curation and selection, not to declare benchmarks (in)valid.

\section{Methodology}
\label{sec:methodology}

To assess whether a benchmark truly evaluates global versus local reasoning, we propose a patch-based evaluation framework. % centered on the Region Comprehension Index (RCI). 
RCI is designed as a practical, model-based score for dataset auditing and curation, not for optimizing model performance. We emphasize that RCI is a descriptive, not prescriptive, metric. It does not label a benchmark as “invalid” because many items can be solved with localized visual cues. Instead, RCI quantifies the type of visual information a benchmark rewards, helping dataset designers align benchmarks with the intended reasoning requirements of their applications. This framing underscores our focus on aiding dataset curation and benchmark evaluation, rather than optimizing specific model performance.

Below, we detail our evaluation framework and formalize RCI.

\subsection{Patch-Based Evaluation Framework}
\label{sec:patch_framework}

In our framework (Figure~\ref{fig:RCIn_framework}), each image is systematically divided into $n \times n$ non-overlapping, equally sized patches, where $n$ controls spatial granularity. For each patch, we independently evaluate the reference-model's performance, thereby isolating the contribution of localized visual information. This approach reveals whether a benchmark can be solved by focusing on specific regions or genuinely requires holistic image understanding.

We adopt a regular grid partitioning to ensure that patch selection is systematic, unbiased, and easy to interpret. This approach avoids the confusion and implementation overhead of object-centric or saliency-based patching, making results more reproducible and conclusions more comparable across benchmarks. Specifically, we study:

\begin{itemize}
    \item $n=1$ (full image): Baseline for RCI
    \item $n=2$ (four patches): Coarse-level for RCI
    \item $n=3$ (nine patches): Fine-level for RCI
\end{itemize}

We explored higher $n$, in selected experiments (see Appendix~\ref{sec:granularity_ablation}), finding diminishing gains \& substantially increased computational cost ($n^2$).

\subsection{Region Comprehension Index (RCI)}

RCI quantifies the extent to which solving a dataset’s tasks requires global versus localized visual information. For patch granularity $n$, RCI is defined as:

\begin{equation}
    \text{RCI}_n = 1 - \frac{\text{MPP}_n}{\text{FIP}}
\end{equation}

where:
\begin{itemize}
    \item $\text{MPP}_n$ (Maximum Patch Performance): The aggregated model performance on the best-performing individual patch (per sample) is used, at granularity $n$.
    \item $\text{FIP}$ (Full Image Performance): The aggregated model performance over full-image for each sample.
\end{itemize}

RCI does not require patch-level annotations. Instead, it reuses existing item labels and selects the best-performing patch per item using the benchmark’s native scorer, ensuring no new annotations are necessary. This makes RCI a model-based audit tool that evaluates dataset reasoning requirement without altering the dataset itself. See Appendix~\ref{sec:appendix_mpp_fip} for detailed definitions \& intuition.

% RCI does not require patch-level annotations. Instead, patches are inference units scored against existing gold labels, ensuring no new annotations are necessary. This makes RCI a model-based audit tool that evaluates dataset reasoning requirement without requiring changes to the dataset itself. See Appendix~\ref{sec:appendix_mpp_fip} for detailed definitions \& intuition.

% \paragraph{Validity Domain \& Chance.}
% We interpret RCI only when full-image performance (FIP) exceeds a dataset-specific chance floor by a small margin (Appendix ~\ref{sec:chance-validity}). 

% Formally, (dataset, model) pair \((d,m)\) is considered valid if \(\mathrm{FIP}(d,m)\ge \mathrm{chance}(d)+\Delta_{\min}\) with \(\Delta_{\min}=\delta=1.0\) percentage point (or \(\max\{\delta,2\,\mathrm{SE}\}\) when CIs are reported). All cells in our study meet this criterion.

\paragraph{Validity Domain \& Chance.}
We interpret RCI only when the full-image performance (FIP) exceeds a dataset-specific chance threshold by a small margin (Appendix~\ref{sec:chance-validity}). Formally, a (dataset, model) pair \((d,m)\) is considered valid if
\[
\mathrm{FIP}(d,m) \ge \mathrm{chance}(d) + \Delta_{\min},
\]
where \(\Delta_{\min} = \delta = 1.0\) percentage point by default, or \(\max\{\delta, 2\,\mathrm{SE}\}\) when confidence intervals (CIs) are reported. All dataset-model pairs in our study satisfy this criterion.

\paragraph{Interpretation:} RCI can be interpreted as shown in Table \ref{tab:rci-interpretation}. RCI is not a metric in the geometric sense, nor is it a normalized score between 0 and 1, but rather a comparative score. As with other model-based evaluation scores (e.g., CLIPScore), its absolute value may vary across datasets and the reference-models used. However, in our experiments, we observe that RCI trends are robust across a variety of reference-model architectures, underscoring its practical utility for both academic and industrial benchmarking.  %For complete mathematical details and formal definitions of MPP$_n$ and FIP, see Appendix~\ref{sec:appendix_mpp_fip}.
% We further study the stability and human alignment of RCI in subsequent sections and in the Appendix.

\begin{table}[!ht]
\centering
\scalebox{0.90}{
\begin{tabular}{p{2.6cm} p{4.0cm}}
\toprule
\textbf{RCI Value} & \textbf{Interpretation} \\
\midrule
$\text{RCI} \gg 0$   & Task requires strong global reasoning \\
$\text{RCI} \approx 0$ & Balanced global and local reasoning \\
$\text{RCI} \ll 0$   & Task can be solved with localized visual cues \\
\bottomrule
\end{tabular}
}
\caption{Interpretation of RCI values in terms of task reasoning requirements.}
\label{tab:rci-interpretation}
\vspace{-1.5em}
\end{table}

\paragraph{Guidance for RCI.}
To facilitate interpretation of RCI values, we introduce qualitative bands based on the dataset's requirement for local vs. global visual information:

\begin{itemize}
    \item \textbf{Strong local} (\( \text{RCI} \leq -0.30\)): The dataset heavily rewards local visual features, with limited global reasoning required for high model performance.
    \item \textbf{Moderate local} (\(-0.30 < \text{RCI} \leq -0.10\)): The dataset relies on local features but still requires some degree of global reasoning.
    \item \textbf{Balanced} (\(-0.10 < \text{RCI} \leq +0.10\)): The dataset requires a balance of local and global reasoning, with no clear preference for one over the other.
    \item \textbf{Moderate global} (\(+0.10 < \text{RCI} \leq +0.30\)): The dataset favors global reasoning but still has elements where local features are important.
    \item \textbf{Strong global} (\( \text{RCI} \geq +0.30\)): The dataset predominantly rewards global visual information, with little reliance on local features.
\end{itemize}

These bands provide a clear mapping of the RCI value to an intuitive understanding of the dataset's requirements for local vs. global visual reasoning. We also recommend percentile-based calibration, especially when applying RCI to larger datasets, where adjustments may be needed depending on the dataset’s inherent structure.

\subsection{Spatial Bias Analysis}

To diagnose whether certain image regions disproportionately affect performance on benchmark , we analyze the contribution of each patch across the dataset. Specifically, for every patch position, we compute the fraction of total correct predictions (or score) resulting from that patch when used in isolation. This reveals if particular regions, such as the image center: dominate task success, indicating potential spatial shortcuts or artifacts in the benchmark design. Identifying such biases helps dataset creators diversify content placement and mitigate unintended model shortcuts.

\section{Experiments and Results}

We apply our RCI-based evaluation on a comprehensive range of benchmarks, ensuring our evaluation captures variations across task  types, visual context granularities and reference-models.

\subsection{Experimental Setup}
\label{sec:experimental_setup}

\paragraph{Reference Models for \( \text{RCI} \)}

To comprehensively understand dataset designs and %evaluate biases and
the dependency on global versus localized reasoning, we selected models based on the following key criteria: (1) Architectural Diversity, (2) Reasoning Capabilities, (3) Grounding and Localization Sensitivity, and (4) Scalability and Efficiency. Based on these we, shortlisted InternVl-2.5-1B, Qwen2-VL-2B, and Molmo-1B models for \( \text{RCI} \) computation.
% are experiments and \( \text{RCI} \) computation. 

% \paragraph{Patch Granularity and Reference Model Selection.}
% For our evaluation, we use patch granularities \(n \in \{2, 3\}\), as these provide a balanced trade-off between computational cost and meaningful insight into the dataset's reasoning requirements. Finer granularity leads to diminishing returns and increased computational overhead. We recommend using these values for efficient yet robust audits, as detailed further in Appendix A.4.3. 

% In terms of model selection, we choose small reference models (e.g., InternVL-2.5-1B, Qwen2-VL-2B, and Molmo-1B), as their performance aligns closely with larger models (r ≥ 0.91), ensuring consistent RCI trends while minimizing inference cost. For a detailed analysis of model selection and granularity effects, see Appendix A.4.1 and A.4.3.

\paragraph{Datasets \& Benchmarks}

We evaluate RCI across the following vision-language benchmarks:

% multiple vision-language benchmarks as follows:

 - Multiple-Choice QA (MCQ): AI2D, BLINK, MMStar, ScienceQA, RealWorldQA.
 
 - Yes/No Classification: AMBER, MME, POPE, HallusionBench.
 
 - Visual Question Answering (VQA): GQA, ChartQA, TextVQA, VizWiz.
 
 % - Image Captioning: MS-COCO Captions

\paragraph{Evaluation Protocol.}
We evaluate using patch granularities \( n \in \{2, 3\} \), balancing computational efficiency with the dataset's reasoning requirements. While finer granularities offer more detail, they incur diminishing returns and increased computational cost. The evaluation is performed using InternVL-2.5-1B, Qwen2-VL-2B, and Molmo-1B models, ensuring robustness across different model types. The smaller models align closely with larger models (r > 0.9) in terms of performance, providing an efficient yet reliable evaluation.  

We recommend using these values for efficient yet robust audits. For further details on evaluation protocol, model selection and granularity, refer to Appendix~\ref{sec:experimental_setup_extended}. The benchmark datasets are further elaborated in Appendix \ref{sec:data_models}.

% We further discuss the details of the evaluation protocol, model selection criteria, and datasets in Appendix \ref{sec:experimental_setup_extended} and \ref{sec:data_models}.

\subsection{Results and Discussion}
\label{sec:results}

We organize our analysis around core research questions central to evaluating RCI’s validity and utility. Each subsection directly addresses one of these questions.

\begin{figure}[!th]
    \centering
    \includegraphics[width=\linewidth]{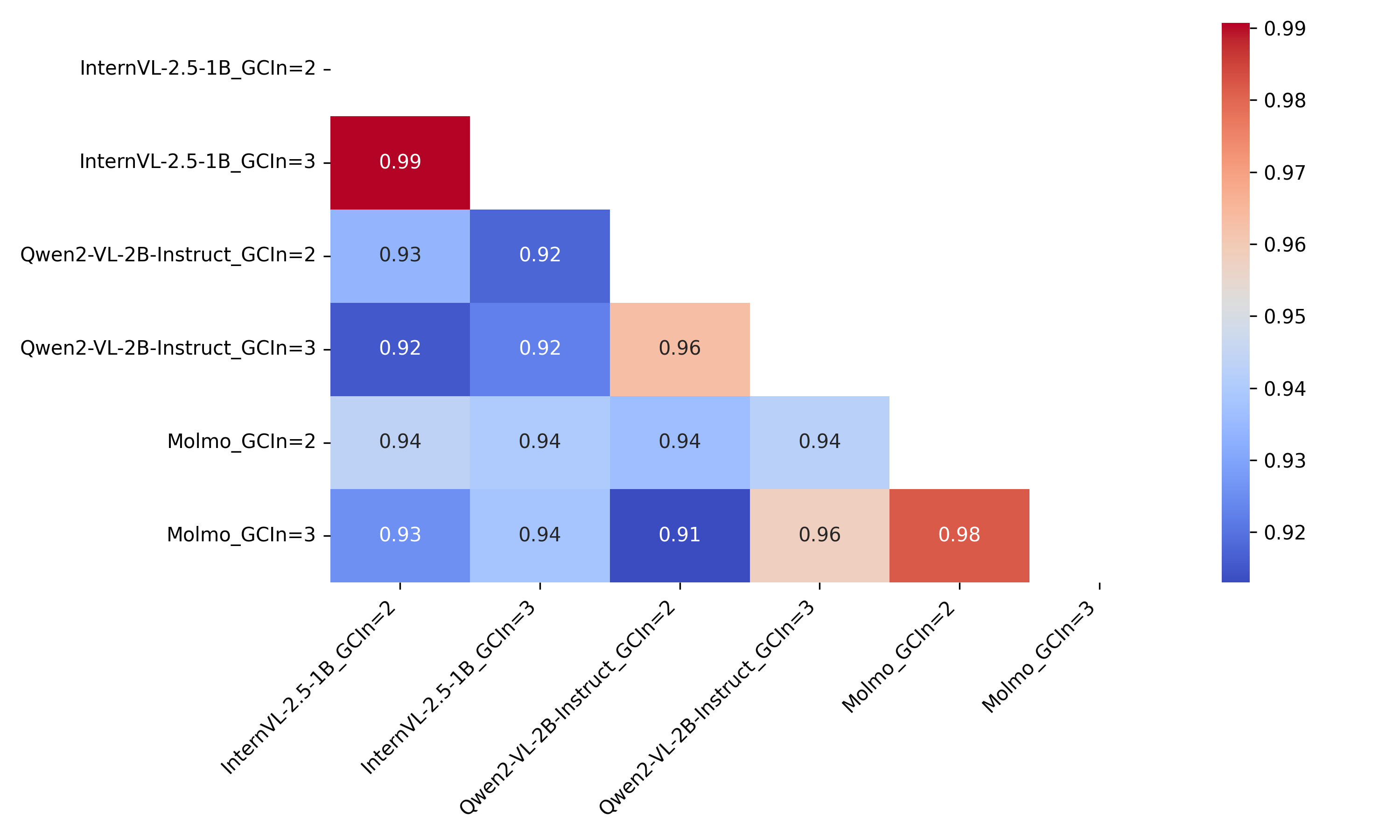}
    \caption{Cross-model correlation of RCI at n=2, 3. High correlations indicates RCI is inherent to dataset rather than specific model behaviors.}
    % that dataset-induced biases are consistent across diverse model architectures,confirming the reliability of \( \text{RCI} \).}
    % InternVL-2.5, Qwen2-VL, and Molmo indicate robust detection of dataset biases.}
    % \caption{Cross-model correlation of RCI scores (n=2,3). High correlations across InternVL-2.5, Qwen2-VL, and Molmo indicate robust detection of dataset biases.}
    \vspace{-1em}
    \label{fig:model_correlation}
\end{figure}

% \subsubsection{Cross-Model Consistency of \( \text{RCI} \)}
 % \subsubsection{\( \text{RCI} \) Consistency across Models}
 \subsubsection{Does the Choice of Reference-Model Affect RCI?}

% ACL
% We evaluate the robustness of RCI across reference-models with diverse architectures (InternVL-2.5, Qwen2-VL, Molmo) and varying scales. As shown in Figure~\ref{fig:model_correlation}, RCI values are highly correlated across all models (\( r \geq 0.91 \)), indicating that the global vs.\ local reasoning dependencies exposed by RCI are intrinsic to the datasets, not artifacts of specific model choices. Further, when comparing small and large variants of these models (e.g., Qwen2-VL-2B vs.\ Qwen2-VL-72B), we observe consistently high intra-family correlations (\( r \geq 0.86 \)), supporting the use of efficient, smaller models for RCI-based dataset audits without sacrificing diagnostic fidelity. Detailed results, correlation matrices, and qualitative analyses are provided in Appendix~\ref{sec:bigger_model_comparison}, reinforcing RCI’s reliability for both academic and industrial settings.

% \subsubsection{Does the Choice of Reference Model Affect RCI?}

We assess the robustness of RCI across diverse reference models and scales, including InternVL-2.5, Qwen2-VL, and Molmo. As shown in Figure~\ref{fig:model_correlation}, RCI values are highly correlated across architectures (\( r \geq 0.91 \)), demonstrating that the global versus local reasoning dependencies measured by RCI are intrinsic to the datasets, rather than artifacts of specific model choices. Comparing small and large model variants (e.g., Qwen2-VL-2B vs.\ Qwen2-VL-72B) yields similarly high intra-family correlations (\( r \geq 0.86 \)), supporting the use of efficient, smaller models for RCI-based dataset audits without loss of diagnostic power. Detailed correlation matrices and further analysis are provided in Appendix~\ref{sec:bigger_model_comparison}, confirming RCI’s reliability for both research and industry applications.

\subsubsection{Does RCI Reflect Human Reasoning in Vision-Language Tasks?}

To further validate RCI, we conducted a small-scale human study in which annotators solved representative benchmark tasks using individual image patches versus full images. Human performance exhibited similar trends as model-based RCI: tasks with high RCI values were consistently more difficult for humans when restricted to local patches, while tasks with low RCI could often be solved from partial information. This alignment suggests RCI not only diagnoses dataset biases for models, but also reflects human reasoning expectations. Further details are in Appendix~\ref{sec:human_study}.

\begin{table*}[!ht]
    \centering
    \renewcommand{\arraystretch}{1.2} % Adjust row height
    \setlength\dashlinedash{0.5pt}    % Dash length
    \setlength\dashlinegap{1pt}       % Gap between dashes
    \setlength\arrayrulewidth{0.8pt}  % Thickness of outer border
    
    % Adjust column width manually instead of using tabularx
    \scalebox{0.85}{ 
    \begin{tabular}{|l|>{\centering\arraybackslash}p{1.8cm}:>{\centering\arraybackslash}p{1.8cm}|%
                       >{\centering\arraybackslash}p{1.8cm}:>{\centering\arraybackslash}p{1.8cm}|%
                       >{\centering\arraybackslash}p{1.8cm}:>{\centering\arraybackslash}p{1.8cm}|}
        \hline
        \textbf{Dataset} & \multicolumn{2}{c|}{\textbf{InternVL-2.5-1B}} & \multicolumn{2}{c|}{\textbf{Qwen2-VL-2B-Instruct}} & \multicolumn{2}{c|}{\textbf{Molmo-1B}} \\
        \cline{2-7}
        & RCI$_{n=2}$ & RCI$_{n=3}$ & RCI$_{n=2}$ & RCI$_{n=3}$ & RCI$_{n=2}$ & RCI$_{n=3}$ \\
        \hline
        AI2D\_TEST & -0.171 & -0.215 & -0.112 & -0.159 & -0.224 & -0.324 \\
        AMBER & -0.010 & -0.028 & -0.013 & -0.027 & -0.015 & -0.038 \\
        BLINK & -0.231 & -0.294 & -0.204 & -0.367 & -0.383 & -0.516 \\
        ChartQA\_TEST & 0.202 & 0.290 & 0.243 & 0.290 & 0.198 & 0.237 \\
        GQA\_TestDev\_Balanced & -0.207 & -0.266 & -0.189 & -0.265 & -0.235 & -0.310 \\
        HallusionBench & -0.267 & -0.346 & -0.223 & -0.353 & -0.216 & -0.355 \\
        MME & -0.064 & -0.097 & -0.107 & -0.117 & -0.134 & -0.171 \\
        MMStar & -0.235 & -0.286 & -0.262 & -0.389 & -0.296 & -0.458 \\
        POPE & -0.054 & -0.055 & -0.055 & -0.068 & -0.044 & -0.051 \\
        RealWorldQA & -0.210 & -0.307 & -0.170 & -0.272 & -0.222 & -0.315 \\
        ScienceQA\_TEST & 0.037 & 0.060 & 0.071 & 0.124 & 0.044 & 0.080 \\
        TextVQA\_VAL & 0.063 & 0.112 & 0.075 & 0.119 & 0.093 & 0.136 \\
        VizWiz & -0.099 & -0.122 & -0.030 & -0.036 & -0.083 & -0.092 \\
        \hline
    \end{tabular}
    }
    \caption{\( \text{RCI} \) scores for 13 multimodal benchmarks across three reference MLLMs and two patch granularities ($n=2, 3$),  highlighting the reasoning requirements of each dataset as measured by RCI.}
    \vspace{-1em}
    \label{tab:model_comparison}
\end{table*}

\begin{figure*}[!ht]
    \centering
    \includegraphics[width=0.95\linewidth]{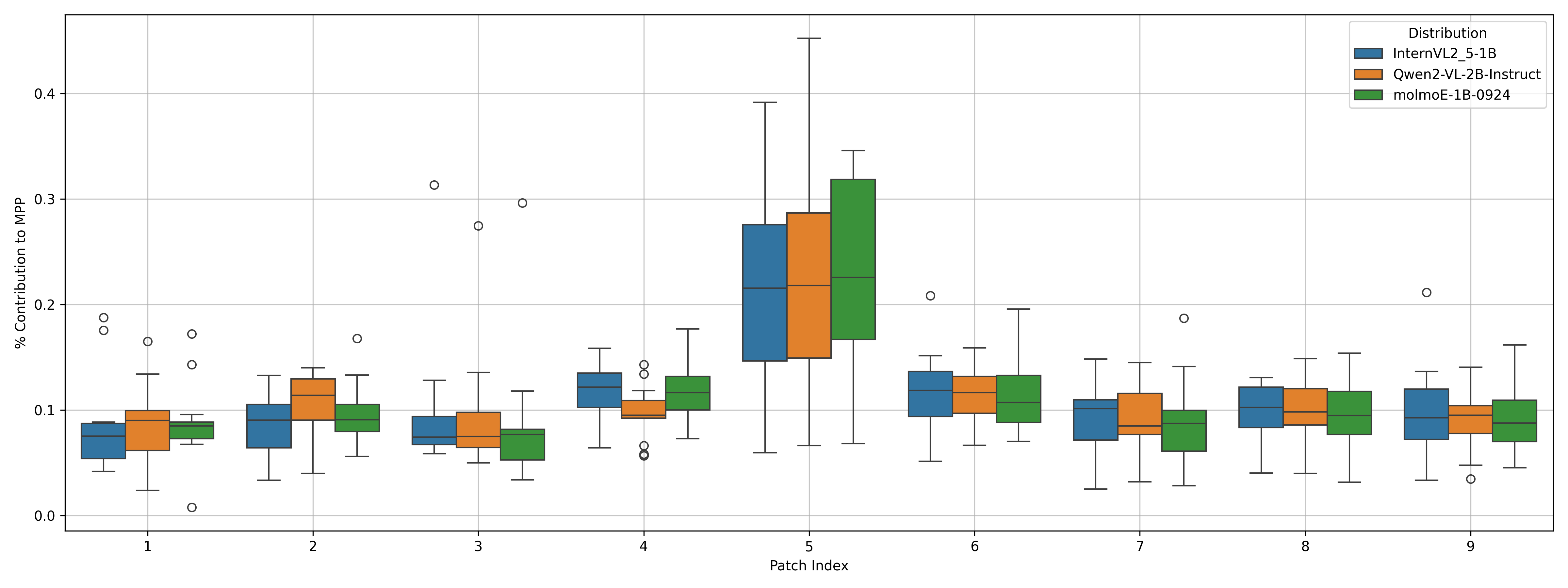}
    \caption{Avg. contribution of each patch to \(\text{MPP}_{n=3}\) across all datasets. Central patches (Patch 5) dominate, highlighting localized biases, whereas peripheral patches contribute minimally.}
    % \caption{Patch-wise \(\text{MPP}_{n=3}\) distributions across all evaluated datasets. The results highlight strong localized spatial dependency, with Patch 5 (Center) contributing the most to model performance. Patch 3 (Top-Right) exhibits the highest variance, indicating dataset-dependent importance, while peripheral and bottom-row patches (1, 7, 8, 9) contribute the least.}
    \vspace{-1em}
    \label{fig:violin_RCIn_3}
\end{figure*}

\begin{figure}[!ht]
    \centering
    \includegraphics[width=\linewidth]{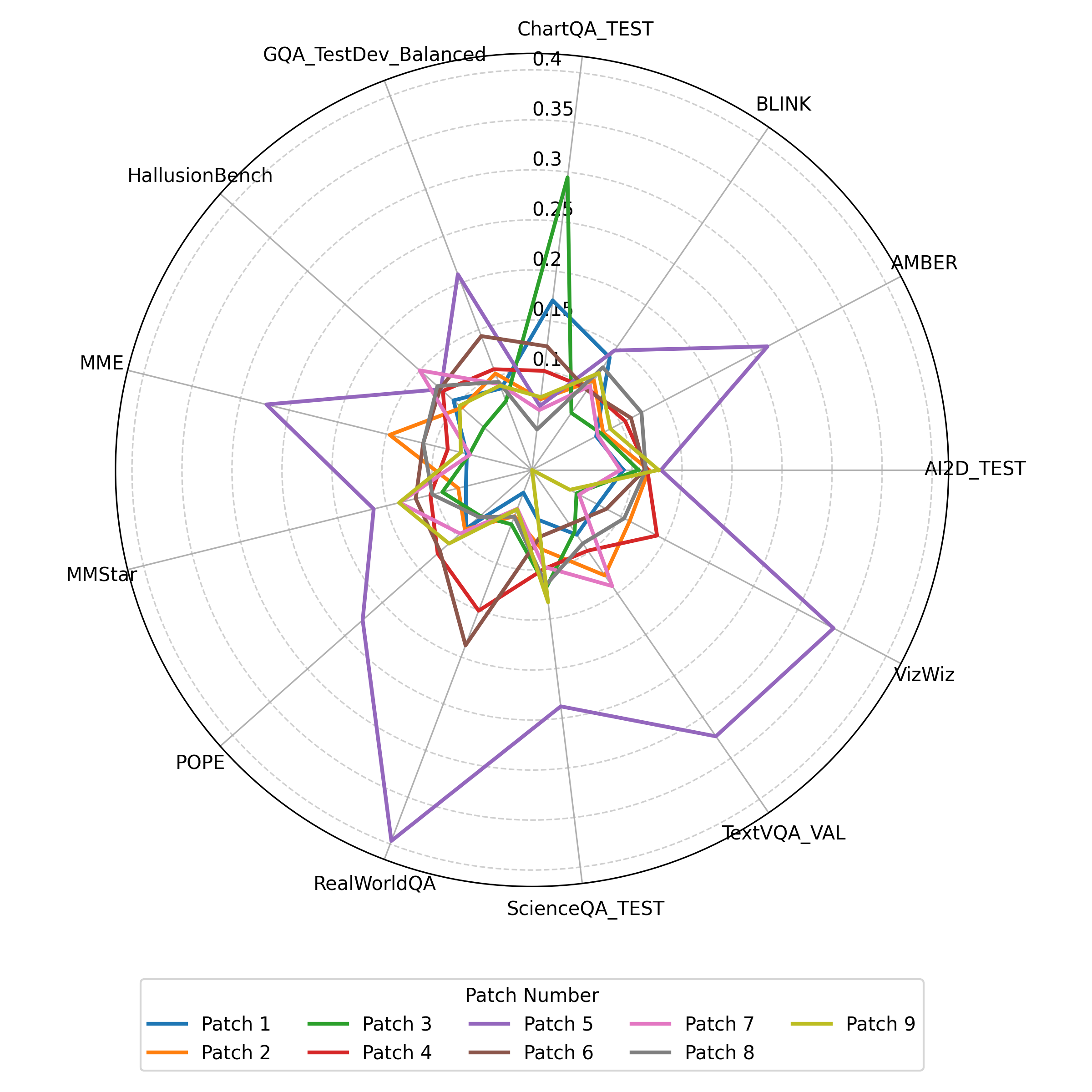}
    \caption{Distribution of each patch patch contribution for \(\text{MPP}_{n=3}\). Patch 5 consistently dominates, reinforcing strong central spatial biases. }
    % \caption{Contribution of each patch for \(\text{MPP}_{n=3}\).The results reinforce localized spatial biases, with Patch 5 (Center) consistently contributing the most to performance. Patch 3 (Top-Right) exhibits dataset-dependent variability, while bottom patches (7, 8, 9) contribute the least, indicating limited reliance on peripheral information.}
    \vspace{-1.5em}
    \label{fig:RCIn_3x3_radar}
\end{figure}

% \subsubsection{RCI Analysis}
\subsubsection{What Reasoning Biases Does RCI Reveal?}

Table~\ref{tab:model_comparison} summarizes \( \text{RCI} \) results across 13 multimodal benchmarks at patch granularities \( n=2,3 \). We identify distinct %dataset 
patterns in terms of dependency on global  reasoning versus localized reasoning.% visual cues.

\paragraph{Benchmarks in Favor of Localized Reasoning.}
% Several benchmarks, notably 
Benchmarks like BLINK, HallusionBench, MMStar, RealWorldQA, GQA, and AI2D, consistently exhibit strong negative \( \text{RCI} \) values. Negative RCI indicates models achieve superior performance using smaller image patches than full images, clearly highlighting substantial dependence on localized reasoning. % biases.
Such benchmarks thus permit solving the tasks using localized visual cues rather than enforcing global reasoning.

\paragraph{Benchmarks in Favor of Global Reasoning.}
Benchmarks like ChartQA, ScienceQA, and TextVQA consistently yield positive or near-neutral RCI values, explicitly indicating their reliance on comprehensive global reasoning. For instance, ChartQA tasks require interpreting multiple visual elements like axes, legends, \& graphical data simultaneously, %effectively 
enforcing integration of distributed visual information rather than localized cues alone.

% \vspace{-1.6em}
\paragraph{Influence of Patch Granularity.}
Decrease in \( \text{RCI} \) values observed at finer granularity (\( n=3 \)) further exposes localized reasoning biases. For example, MMStar RCI drops notably from -0.235 (\( n=2 \)) to -0.286 (\( n=3 \)) for \( \text{RCI} \) with InternVL-2.5, and more drastically for \( \text{RCI} \) with Molmo from -0.296 to -0.458. This confirms that finer granularity patches accentuate dataset biases toward localized reasoning, highlighting the inadequacy of existing benchmarks to robustly assess global reasoning. We discuss higher values of $n$, model-specific trends \& illustrative examples in Appendix \ref{sec:granularity_ablation}, %\ref{sec:model_wise_extended}
\& \ref{sec:example_RCI}.
% provide some illustrative examples in Appendix \ref{sec:example_RCI}

% \subsection{Spatial Bias Analysis}
\subsubsection{Do Benchmarks Exhibit Systematic Spatial Biases?}
\label{sec:spatial_bias}
We further analyze spatial biases, highlighting how visual information is disproportionately distributed across image patches within benchmarks.

\paragraph{Central Spatial Bias (Patch 5).}
Figure~\ref{fig:violin_RCIn_3} and \ref{fig:RCIn_3x3_radar} clearly indicates a strong central bias across benchmarks, with the center patch (Patch 5) consistently contributing the highest to model performance. This confirms a pervasive dataset design flaw where central visual information disproportionately influences performance, promoting localized reasoning rather than holistic image understanding.

\paragraph{Peripheral Underrepresentation.}
Peripheral patches (like 1, 3, 7, and 8) consistently exhibit minimal contributions, suggesting benchmarks rarely include significant information in these areas. Such biases hinder the assessment of comprehensive spatial reasoning, potentially compromising model robustness.

\paragraph{Dataset-specific Variability.}
Certain benchmarks exhibit notable variance in patch contributions. For instance, ChartQA demonstrates substantial reliance on Patch 3 (top-right) (Figure \ref{fig:RCIn_3x3_radar}), reflecting dataset-specific layouts such as chart legends or annotations. This emphasizes the importance of explicitly understanding dataset structures to avoid inadvertent biases.

\paragraph{Patch Granularity Sensitivity.}
Comparing results at \( n=2 \) and \( n=3 \) granularities (Figure~\ref{fig:heatmap_patch_3}, ~\ref{fig:heatmap_patch_2}) we observe finer patches (\( n=3 \)) clearly expose spatial biases more prominently than coarser patches (\( n=2 \)). Larger patches naturally cover critical regions more evenly, masking biases, while smaller patches reveal precise localization biases. Despite computational constraints, the choice of \( n=3 \) strikes a balance between detailed analysis and computational efficiency, effectively capturing meaningful spatial biases.

All reported results fall within the validity domain: for every dataset–model pair, FIP exceeds the dataset-specific chance floor by at least \(\Delta_{\min}\) (Appendix~\ref{sec:chance-validity}).

\section{Practical Implications for Industry and Deployment}
\label{sec:discussion}

RCI provides an actionable lens for aligning multimodal datasets and benchmarks with the specific reasoning needs of real-world applications. By quantifying the dependency on global versus local visual information, RCI enables practitioners to:

\begin{itemize}
    \item \textbf{Audit and select datasets} based on the reasoning requirements of target deployment scenarios (e.g., favoring high-RCI datasets for applications needing holistic understanding).
    \item \textbf{Detect and address spatial biases} during dataset development, supporting the creation of more balanced and robust benchmarks.
    \item \textbf{Continuously monitor and maintain} the alignment between production data, benchmark tasks, and application demands as systems evolve.
\end{itemize}

For detailed practitioner guidance and application examples, see Appendix~\ref{sec:practitioner_appendix}. Further discussion of limitations and future directions is provided in Appendix~\ref{sec:limitations}.

\section{Conclusion}
\label{sec:conclusion}

We introduced the Region Comprehension Index (RCI), the first model-based score to explicitly quantify whether multimodal benchmarks are suitable for global or local visual reasoning to solve tasks. By providing a transparent measure of spatial reasoning dependency, RCI empowers researchers and practitioners to curate and evaluate datasets that are better aligned with real-world application needs, supporting the development of robust MLLMs for deployed systems.

Through systematic evaluation on 13 widely-used benchmarks, we found that most current datasets favor localized reasoning and often contain significant spatial biases and shortcuts, potentially undermining model robustness and generalization in practical settings.

RCI is already integrated into enterprise workflows to guide dataset construction and evaluation to train production-grade multimodal models. Looking forward, we will extend RCI to sequential and broader multimodal domains such as video and audio, further enabling reliable, production-ready AI systems.

% ACL Industry Track
% \section{Conclusion}
% \label{sec:conclusion}

% We introduced the Region Comprehension Index (RCI), a metric explicitly designed to quantify multimodal datasets and benchmarks dependencies on global versus localized visual reasoning. By evaluating on 13 widely-used benchmarks, RCI identified significant dataset-induced biases toward localized reasoning, potentially misleading perceptions of multimodal progress and risking practical failures in tasks requiring robust global context reasoning.

% Integrating RCI into benchmark evaluation and dataset curation, researchers and practitioners can explicitly quantify and address spatial reasoning biases, creating datasets aligned with practical reasoning requirements. Future research directions may include extending RCI to sequential and other multimodal domains (e.g., video and audio), further enriching our understanding and facilitating robust real-world multimodal AI deployments.

\section*{Limitations}

While RCI is model-based, our experiments show that its trends are robust across diverse open-source architectures and scales. Nevertheless, absolute RCI values may still reflect reference model choices, and future work could explore ensembles or standardized model sets for evaluation. Our analysis is limited to visual benchmarks and does not address temporal or audio modalities, which present new challenges for reasoning assessment. 

Additionally, RCI computation requires multiple model inferences per image (for $n=2,3$ patch granularities), leading to increased cost on large datasets; efficient sampling or approximation methods are a promising area for further research. Our study focuses on open-source models due to cost and reproducibility; extending RCI to proprietary or closed-source systems remains an open question.

Expanded discussion of these points and further future directions can be found in Appendix~\ref{sec:limitations}.

% % % Bibliography entries for the entire Anthology, followed by custom entries
% % %\bibliography{anthology,custom}
% % % Custom bibliography entries only
\bibliography{custom}

\clearpage

\appendix

\section{Appendix}
\label{sec:appendix}

\subsection{Expanded Definitions: MPP and FIP}
\label{sec:appendix_mpp_fip}

RCI is based on two key quantities, each computed using the native evaluation metric (e.g., accuracy, CIDEr, VQA score) of the benchmark:

\textbf{Maximum Patch Performance (MPP$_n$):}  
For patch granularity $n$, each image is divided into $n \times n$ non-overlapping patches. For each image-question pair, we compute the model’s prediction on each patch and select the highest-performing patch as the prediction for that sample. The set of all such “best-patch” predictions is then used to compute the official benchmark metric to yield MPP$_n$:
\[
\text{MPP}_n = \mathcal{M}\big(\{\max_{p \in \mathcal{P}_i^{(n)}} F(p, q_i)\}_{i=1}^N\big)
\]
where $\mathcal{M}$ is the benchmark’s evaluation metric (e.g., accuracy), $\mathcal{P}_i^{(n)}$ is the set of $n \times n$ patches for image $i$, $q_i$ is the question or prompt, $F$ is the reference model’s prediction for patch $p$ and query $q_i$, and $N$ is the dataset size.

\textbf{Full Image Performance (FIP):}  
For the same dataset, FIP is the model’s performance when using the full image for each input:
\[
\text{FIP} = \mathcal{M}\big(\{F(\text{img}_i, q_i)\}_{i=1}^N\big)
\]
where $\text{img}_i$ is the original image.

\textbf{Key Property:}  
Since both MPP$_n$ and FIP use the same data and evaluation metric, RCI is invariant to metric choice and directly comparable across benchmarks.

\textbf{Intuitive Explanation:}  

- MPP$_n$: “What is the best the model can do if it only sees one patch of the image?”

- FIP: “How well does the model do with the whole image?”

- If MPP$_n$ is much less than FIP, the tasks in the dataset needs information from multiple regions (global reasoning).

- If MPP$_n$ is close or more to FIP, the tasks in the dataset can be solved well with individual patches (localized reasoning).

% \subsection{Chance \& Validity Checks}
% \textbf{Chance floors.} We define dataset-specific chance baselines consistent with the dataset’s official scorer:
% (i) \textit{Multiple-choice QA}: uniform guess \(1/|C|\) over options; 
% (ii) \textit{Yes/No}: 0.5; when label skew is material, we additionally compute a majority-class baseline and use \(\max(0.5,\text{majority})\);
% (iii) \textit{Open-ended VQA/short-answer}: the majority-answer baseline evaluated with the official scorer (for VQA-style consensus metrics this yields a near-zero floor).

% \noindent\textbf{Validity rule.} A cell is valid if \(\mathrm{FIP} \ge \mathrm{chance} + \delta\), with \(\delta=1.0\) pp by default (or \(\ge 2\times\)SE when CIs are available). 

% \noindent\textbf{Finding.} All dataset–model pairs in this paper meet the criterion; thus no cells were suppressed.

\subsection{Chance Floors \& Validity Domain}
\label{sec:chance-validity}

\paragraph{Chance floors.}
We define a dataset-specific chance baseline, consistent with each dataset’s official scorer:
\begin{itemize}
\item \textit{Multiple-choice QA}: uniform guess \(1/|C|\) over options.
\item \textit{Yes/No}: \(0.5\); if label skew is material, we also compute the majority-class baseline and use \(\max(0.5,\ \text{majority})\).
\item \textit{Open-ended VQA / short-answer}: the majority-answer baseline evaluated with the official scorer (for VQA-style consensus metrics this is typically near zero).
\end{itemize}

\paragraph{Validity rule.}
We interpret RCI only in non-degenerate regimes where full-image performance (FIP) exceeds the dataset’s chance floor by a small absolute margin. A (dataset, model) cell \((d,m)\) is deemed \emph{valid} if
\begin{equation}
\label{eq:validity}
\mathrm{FIP}(d,m) \;\ge\; \mathrm{chance}(d) \;+\; \Delta_{\min},
\end{equation}
with \(\Delta_{\min}=\delta\) and \(\delta=1.0\) percentage point by default. When confidence intervals are reported, we adopt a more conservative threshold \(\Delta_{\min}=\max\{\delta,\ 2\,\mathrm{SE}\}\), where \(\mathrm{SE}\) is the standard error estimated over items (nonparametric bootstrap by default).

\paragraph{Outcome.}
All dataset–model pairs in this paper satisfy~\eqref{eq:validity}; consequently, no pairs are suppressed and all reported RCI values are interpreted. The datasets categorization are provided in Table~\ref{tab:chance}.

\subsection{Extended Experimental Setup}
\label{sec:experimental_setup_extended}

\paragraph{Benchmark Evaluation Procedure.}
To systematically quantify reasoning dependencies within benchmarks, we follow a structured evaluation procedure: 

\begin{enumerate} 
\item Obtain model predictions for each image patch independently. \item Identify the patch achieving the highest accuracy relative to ground-truth labels. 
\item Compute the Maximum Patch Performance \(\text{MPP}_n\), representing the best achievable performance using a single patch at granularity \( n \). 
\item Calculate the Full Image Performance \(\text{FIP}\) by evaluating models on full, unaltered images. 
\item Derive \(\text{RCI}\) using 
\(\text{MPP}_n\)  and \(\text{FIP}\) for the specified granularity. 

\end{enumerate}
Each evaluation is executed three times, and we report the averaged results, observing negligible variance across runs. For consistency and reproducibility, evaluations are conducted using VLMEvalKit \cite{duan2024vlmevalkit}.

% To quantify reasoning dependencies clearly, we:
% \begin{enumerate}
%     \item Infer the prediction of the model on each patch individually.
%     % \item Evaluate model performance on each patch individually.
%     \item Select the patch which gives the correct prediction as compared to ground-truth.
%     \item Compute the Maximum Patch Performance (\(\text{MPP}_n\)), representing the highest single-patch performance achievable at granularity \( n \).
%     \item Measure the Full Image Performance (\(\text{FIP}\)) using unaltered images.
%     \item Compute \( \text{RCI} \) for the given granularity level.
% \end{enumerate}

% Run the entire evaluation 3 times and report the average. We observed no difference between the runs. We also use VLMEvalKit to run the evaluation against the datasets in this study.

% Our evaluation framework involves the following steps:
% \begin{enumerate}
%     \item \textbf{Patch-Based Evaluation:} Each image is divided into non-overlapping patches with varying granularities (\( n = 2, 3 \)). Models are evaluated on each patch independently to compute patch-level performance.
%     \item \textbf{\( \text{RCI} \) Calculation:} The Region Comprehension Index (\( \text{RCI} \)) is calculated as described in Section~\ref{sec:methodology}, comparing the maximum patch performance (\( \text{MPP}_n \)) with the whole image performance (\( \text{FIP} \)).
%     \item \textbf{Spatial Bias Analysis:} Heatmaps are generated for each \( n \) to visualize spatial biases and identify regions where models rely heavily on localized information.
% \end{enumerate}

\paragraph{Model Selection Criteria}
To comprehensively evaluate dataset biases and the dependency on global versus localized reasoning, we selected models based on the following key criteria:

\begin{itemize}
    \item \textbf{Architectural Diversity:} We include models with varied architectures, ensuring that our findings are not specific to a single model design but instead generalized across different model families.
    \item \textbf{Reasoning Capabilities:} Models are chosen to represent a balance between fine-grained local reasoning and global reasoning. This allows us to assess whether benchmarks promote or hinder global reasoning.
    \item \textbf{Grounding and Localization Sensitivity:} Since \( \text{RCI} \) specifically analyzes spatial bias, models that excel in object grounding, pointing, and spatial reasoning tasks provide deeper insights into patch-level performance.
    \item \textbf{Scalability and Efficiency:} We include models optimized for both efficiency and performance while reducing compute requirements for stable \( \text{RCI} \). 
    % to evaluate whether dataset biases persist across model sizes and computational constraints.
\end{itemize}

\paragraph{Models Used}
Based on these criteria, we selected the following MLLMs for evaluation:

\begin{table*}[t]
\centering
% \small
\scalebox{1.0}{
\begin{tabular}{l l l}
\toprule
\textbf{Task type} & \textbf{Datasets (examples)} & \textbf{Chance(d) definition} \\
\midrule
MCQ & AI2D, BLINK, ScienceQA & \(1/|C|\) (uniform over options) \\
Yes/No & AMBER, MME, POPE, HallusionBench & 0.5 (or majority, whichever is larger) \\
Open-ended & GQA, ChartQA, TextVQA, VizWiz & Majority-answer under official scorer \\
\bottomrule
\end{tabular}}
\caption{Task-aware chance definitions used for validity checks. All cells in our study pass $FIP \ge chance + \delta$.}
\label{tab:chance}
\end{table*}

\paragraph{InternVL 2.5-1B}  \cite{chen2024far,gao2024mini,chen2024expanding,chen2024internvl}
is a large-scale, open-source MLLM that has demonstrated strong performance across diverse multimodal reasoning tasks, including document understanding, commonsense reasoning, and hallucination detection. Its robustness on vision-language benchmarks, including MMMU, makes it particularly suitable for analyzing dataset biases related to local versus global reasoning.

\paragraph{Qwen2-VL-2B} \cite{wang2024qwen2}
 is a lightweight yet powerful MLLM designed for efficient deployment without sacrificing reasoning capabilities. Notably, it exhibits strong performance in multi-image understanding and long-context reasoning, making it an essential candidate for evaluating \( \text{RCI} \), particularly for tasks that require extensive global context comprehension.

\paragraph{Molmo-1B} \cite{deitke2024molmopixmoopenweights}
is an MLLM specifically designed with an emphasis on fine-grained image-text reasoning, grounding, and pointing tasks. Built from scratch using open datasets such as PixMo, Molmo’s architecture is inherently sensitive to spatial biases. Its ability to highlight localized versus global attention patterns makes it a valuable model for our patch-based evaluation and heatmap analyses.

\subsection{Dataset \& Benchmarks}
\label{sec:data_models}

We evaluate our approach across 13 widely-used vision-language benchmarks, covering a diverse range of tasks. These datasets were selected to represent a balanced mix of localized perception tasks (e.g., object recognition) and global contextual reasoning challenges (e.g., complex multiple choice question answering).

\begin{itemize}

    \item \textbf{Visual Question Answering (VQA):}  
    Benchmarks such as GQA \cite{lu2022learnexplainmultimodalreasoning}, ChartQA \cite{masry2022chartqabenchmarkquestionanswering}, TextVQA \cite{singh2019vqamodelsread}, and VizWiz \cite{gurari2018vizwizgrandchallengeanswering} are open-ended VQA tasks where MLLMs must generate responses without restricted answer choices. These benchmarks assess a model's ability to infer answers based on both localized and global scene information.

\begin{figure*}[!ht]
    \centering
    \includegraphics[width=\linewidth]{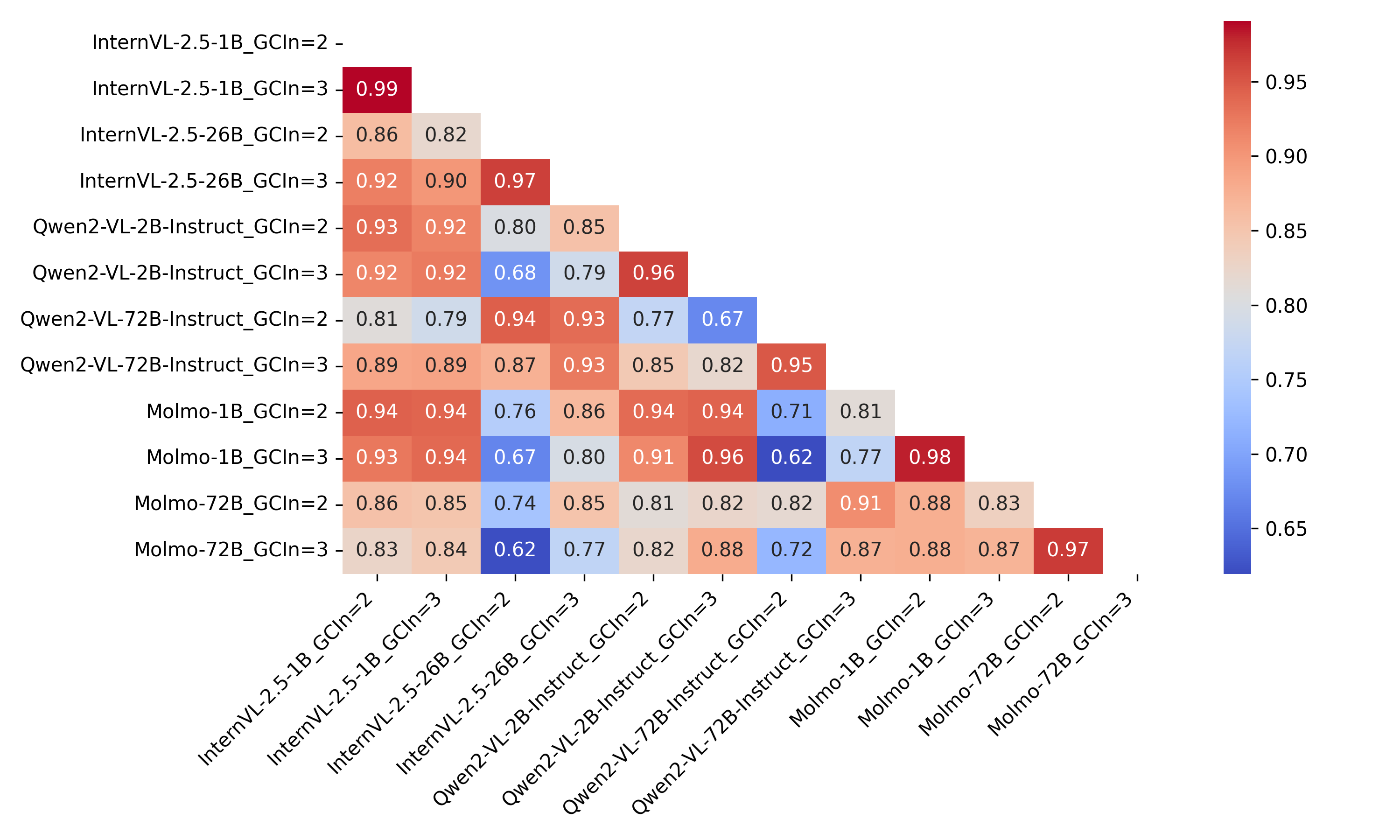}
    \caption{Correlation heatmap comparing \( \text{RCI} \) for n=2,3; between smaller models and their larger-scale counterparts. The results show consistently high intra-model correlations (\(\geq 0.86\)). While larger models show slightly lower cross-model correlation, the findings reinforce that dataset-induced biases persist even in state-of-the-art architectures, validating the use of smaller models for \( \text{RCI} \) evaluation.}
    \label{fig:bigger_model_correlation}
\end{figure*}

\begin{table*}[!ht]
    \centering
    \scalebox{0.90}{
    \renewcommand{\arraystretch}{1.1}
    \begin{tabular}{|l|c|c|c|c|}
        \hline
        \textbf{Dataset} & \textbf{Full Img Perf. (\%)} & \textbf{Patch Perf. (\%)} & \textbf{Perf. Change (\%)} & \textbf{RCI$_{n=3}$ w/ Molmo-1B} \\
        \hline
        ChartQA\_TEST & 93.0 & 66.0 & -27.0 & 0.237 \\
        AMBER & 80.0 & 82.0 & +2.0 & -0.038 \\
        BLINK & 46.0 & 73.0 & +48.0 & -0.516 \\
        \hline
    \end{tabular}}
    \caption{Human accuracy with full images vs. best patch on three benchmarks with varying RCI.}
    \label{tab:human_study}
\end{table*}

    \item \textbf{Multiple-Choice Question Answering (MCQ):}  
    Benchmarks including BLINK \cite{fu2024blinkmultimodallargelanguage}, RealWorldQA \cite{RealWorldQA}, AI2D \cite{kembhavi2016diagramworthdozenimages}, ScienceQA \cite{lu2022learnexplainmultimodalreasoning}, and MMStar \cite{chen2024rightwayevaluatinglarge} provide multiple answer choices, requiring MLLMs to differentiate between options and select the most accurate response. These tasks evaluate multimodal reasoning and answer disambiguation, offering insights into whether datasets support global contextual reasoning.

    \item \textbf{Yes/No Questions (Binary Classification):}  
    Datasets such as POPE \cite{li2023evaluatingobjecthallucinationlarge}, HallusionBench \cite{guan2024hallusionbenchadvanceddiagnosticsuite}, AMBER \cite{wang2024amberllmfreemultidimensionalbenchmark}, and MME \cite{fu2024mmecomprehensiveevaluationbenchmark} focus on binary (yes/no) questions. AMBER, in particular, tests the ability of the model to capture fine-grained spatial relationships, making it useful for evaluating whether a dataset enforces strict spatial comprehension.

    % \item \textbf{Visual Question Answering (VQA):}  
    % Benchmarks such as GQA, ChartQA, TextVQA, and VizWiz are open-ended VQA tasks where MLLMs must generate responses without restricted answer choices. These benchmarks assess a model's ability to infer answers based on both localized and global scene information.

    % \item \textbf{Multiple-Choice Question Answering (MCQ):}  
    % Benchmarks including BLINK, RealWorldQA, AI2D, ScienceQA, and MMStar provide multiple answer choices, requiring MLLMs to differentiate between options and select the most accurate response. These tasks evaluate multimodal reasoning and answer disambiguation, offering insights into whether datasets support global contextual reasoning.

    % \item \textbf{Yes/No Questions (Binary Classification):}  
    % Datasets such as POPE, HallusionBench, AMBER, and MME focus on binary (yes/no) questions. AMBER, in particular, tests a model’s ability to capture fine-grained spatial relationships, making it useful for evaluating whether a dataset enforces strict spatial comprehension.

    % \item \textbf{Image Captioning and Semantic Understanding:}  
    % We include MS-COCO (COCO Captions) to evaluate semantic understanding at the global level. Captioning tasks assess whether models can generate accurate textual descriptions based on an entire image rather than relying on isolated object-level information.
\end{itemize}

\noindent Our dataset selection ensures a diverse evaluation landscape, encompassing tasks that test both localized visual understanding and complex multi-step reasoning.

\subsection{Extended Result \& Discussion}
\label{sec:extended}

\subsubsection{Reference Model and Scale Comparison}
\label{sec:bigger_model_comparison}

To ensure the generalizability of RCI trends, we compare its behavior across both model architectures and scales. As visualized in Figure~\ref{fig:bigger_model_correlation}, high intra-model and intra-family correlations (\( r \geq 0.86 \)) persist even as model size increases (e.g., Qwen2-VL-2B vs.\ Qwen2-VL-72B). Notably, even larger models remain highly correlated, their cross-model correlations are slightly lower. Molmo-72B shows strong alignment with InternVL-2.5-26B and Qwen2-72B, reinforcing that even the largest models, remain highly sensitive to dataset-induced spatial biases, underlining that benchmark improvements must focus on data construction rather than model architecture and scaling alone.

\subsubsection{Human Study: Protocol and Findings}
\label{sec:human_study}

\paragraph{Protocol.}
To assess the alignment between RCI and human reasoning, we selected three representative benchmarks: ChartQA\_TEST (high RCI), AMBER (balanced RCI), and BLINK (low RCI). For each, we sampled approximately 20\% of the dataset (ChartQA\_TEST, AMBER and BLINK). %(ChartQA\_TEST: 100 samples, AMBER: 40, BLINK: 60). 
Three annotators (authors of the paper) answered benchmark questions under two conditions: (1) seeing only the best-performing patch, and (2) seeing the full image. Annotators were not shown model outputs or dataset labels and could consult task instructions as needed. We measured performance in both conditions and collected qualitative feedback on difficulty and task confidence.

\paragraph{Findings.}
Table~\ref{tab:human_study} summarizes the results. For ChartQA\_TEST (high RCI), human- performance dropped sharply from 93\% (full image) to 66\% (patch only), closely mirroring the RCI score of 0.29. For AMBER (balanced RCI), the performance increase was modest (80\% to 82\%), consistent with near-zero RCI. For BLINK (low RCI), humans performed well with the patch-only input, consistent with strong localized cues indicated by negative RCI.

% For BLINK (low RCI), humans performed well with the full image but accuracy dropped significantly to 36\% with patch-only input, consistent with strong localized cues indicated by negative RCI.

Qualitative feedback further supports these trends: annotators reported greater uncertainty and frustration for high-RCI tasks when restricted to patches, noting that key information was often missing. For low-RCI tasks, patch-based answers felt easier and more “guessable,” consistent with shortcut cues.

\paragraph{Conclusion.}
These results demonstrate strong alignment between human performance drops and RCI scores across benchmarks. Thus, RCI not only reveals model-dependent spatial biases, but also aligns with human reasoning demands, supporting its use as a diagnostic tool for dataset and benchmark development.

\subsubsection{Effect of Higher Patch Granularity (\( n=4, 5 \))}
\label{sec:granularity_ablation}

\begin{table}[!ht]
    \centering
    \scalebox{0.90}{
    \renewcommand{\arraystretch}{1.1}
    \begin{tabular}{|l|c|c|c|}
        \hline
        \textbf{Dataset} & \textbf{RCI$_{n=3}$} & \textbf{RCI$_{n=4}$} & \textbf{RCI$_{n=5}$} \\
        \hline
        ChartQA\_TEST & 0.237 & 0.30 & 1.00 \\
        AMBER & -0.038 & -0.030 & 0.55 \\
        BLINK & -0.516 & -0.558 & 0.38 \\
        \hline
    \end{tabular}}
    \caption{RCI values (w/ Molmo-1B) at increasing patch granularity (\( n=3,4,5 \)) for representative benchmarks. Higher $n$ does not yield new trends and can fragment meaningful visual content.}
    \label{tab:rci_granularity}
\end{table}

To evaluate the effect of finer spatial partitioning, we computed RCI values at higher patch granularities ($n=4$ and $n=5$) for three representative benchmarks (ChartQA\_TEST, AMBER, BLINK) using Molmo-1B as the reference model. Results are shown in Table~\ref{tab:rci_granularity}.

We observe that RCI remains relatively stable from $n=3$ to $n=4$, but increases sharply for $n=5$ across all datasets—most notably, RCI$_{n=5}$ for ChartQA\_TEST rises to 1.00, and for AMBER and BLINK to 0.55 and 0.38, respectively. This sharp increase is not indicative of greater global reasoning requirements, but rather reflects an \textit{artificial inflation} caused by excessive patch fragmentation: when each patch contains as little as 4\% of the original image, essential visual context required to solve the task is often lost, dramatically reducing maximum patch performance (MPP) while leaving full-image performance unchanged. This, in turn, inflates the RCI score and obscures the meaningful distinction between local and global reasoning.

These findings highlight a fundamental limitation of overly fine partitioning: as $n$ increases, patches may no longer retain enough semantic information to meaningfully assess reasoning dependencies. Thus, although moderate changes in $n$ (from 2 to 3 or 4) provide some additional insight, further partitioning risks distorting the interpretability and utility of RCI. Based on this analysis, we recommend using $n=2$ or $n=3$ as a practical and meaningful range for spatial bias assessment in most vision-language benchmarks.

% REMOVING AS PER KARAN'S INPUT
% \subsubsection{Model-Specific RCI Analysis}
% \label{sec:model_wise_extended}

% Despite architectural differences, all models display similar spatial dependency trends, confirming that RCI captures dataset-driven, rather than model-specific, reasoning requirements.

\begin{figure*}[!ht]
    \centering
    \includegraphics[width=0.95\linewidth]{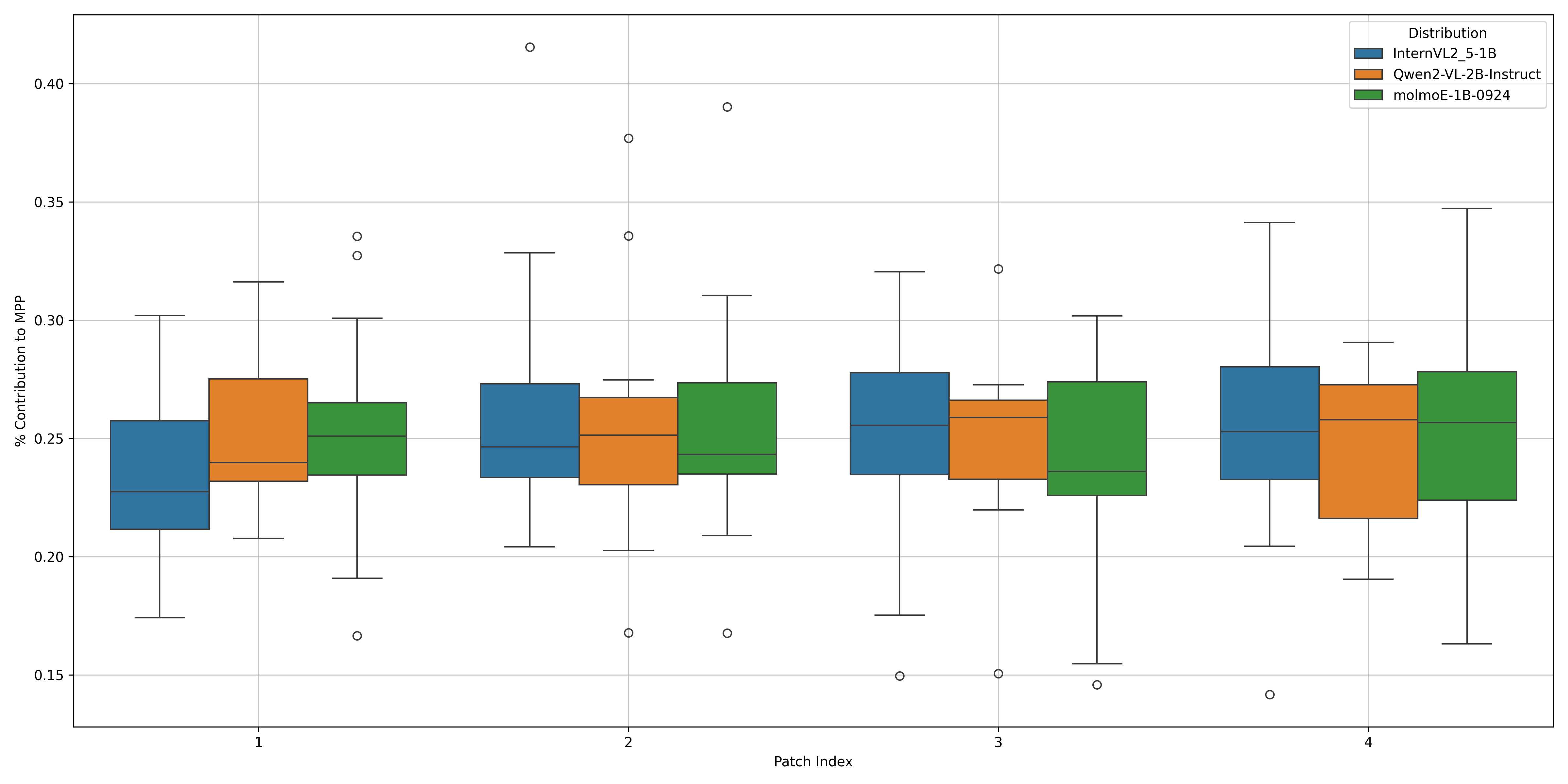}
    \caption{Patch-wise \(\text{MPP}_{n=2}\) distributions across all datasets. Due to larger patches compared to  \(\text{MPP}_{n=3}\), it is harder to isolate localized reasoning dependencies and spatial biases.
    % The results show that localized spatial dependencies are reduced compared to  \(\text{MPP}_{n=3\) with larger patches. 
    Patch 2 exhibits slightly higher variance, suggesting dataset-dependent reliance.}
    \label{fig:violin_RCIn_2}
\end{figure*}

\subsection{Spatial Bias Visualizations and Interpretation}
\label{sec:spatial_bias_appendix}

To complement the quantitative spatial bias analysis in the main text, we present detailed visualizations of patch-wise contributions for all benchmarks and model variants. These analyses provide an interpretable diagnostic for identifying spatial shortcuts, dataset artifacts, and potential weaknesses in benchmark design.

\begin{figure*}[!ht]
    \centering
    \includegraphics[width=\linewidth]{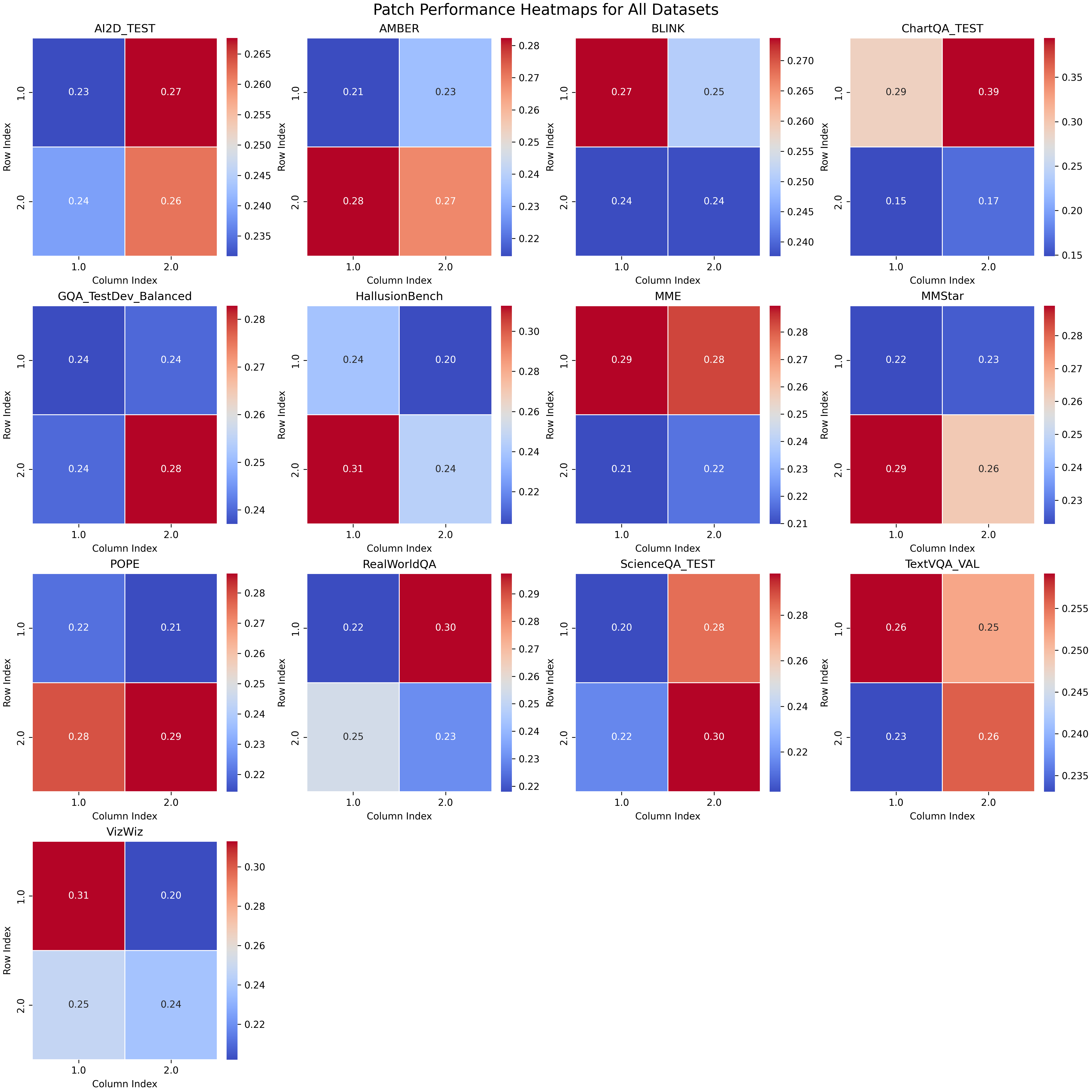}
    % \caption{Patch-wise \(\text{MPP}_{n=2}\) performance heatmaps across datasets. The results show that performance is relatively uniform across patches, but biases remain evident, as models still prioritize certain patches, reinforcing a reliance on localized rather than holistic global image comprehension.}
    \caption{Patch-wise \(\text{MPP}_{n=2}\) performance heatmaps across datasets. The results show that performance is relatively uniform across patches, but  models still prioritize certain patches, reinforcing a reliance on localized rather than global contextual reasoning.}
    \label{fig:heatmap_patch_2}
\end{figure*}

\paragraph{Patch Contribution Distributions (\( n=2 \)).}
Figure~\ref{fig:violin_RCIn_2} (box plot) summarizes the distribution of each patch’s contribution (as a percentage of Maximum Patch Performance, MPP) across all datasets for \( n=2 \) patch granularity and three different model architectures. Uniform distributions would indicate no spatial bias, while skewed contributions signal over-reliance on certain regions. We observe that, on average, central and lower-right patches contribute more, suggesting that many datasets embed critical information non-uniformly.

\begin{figure*}[!ht]
    \centering
    \includegraphics[width=\linewidth]{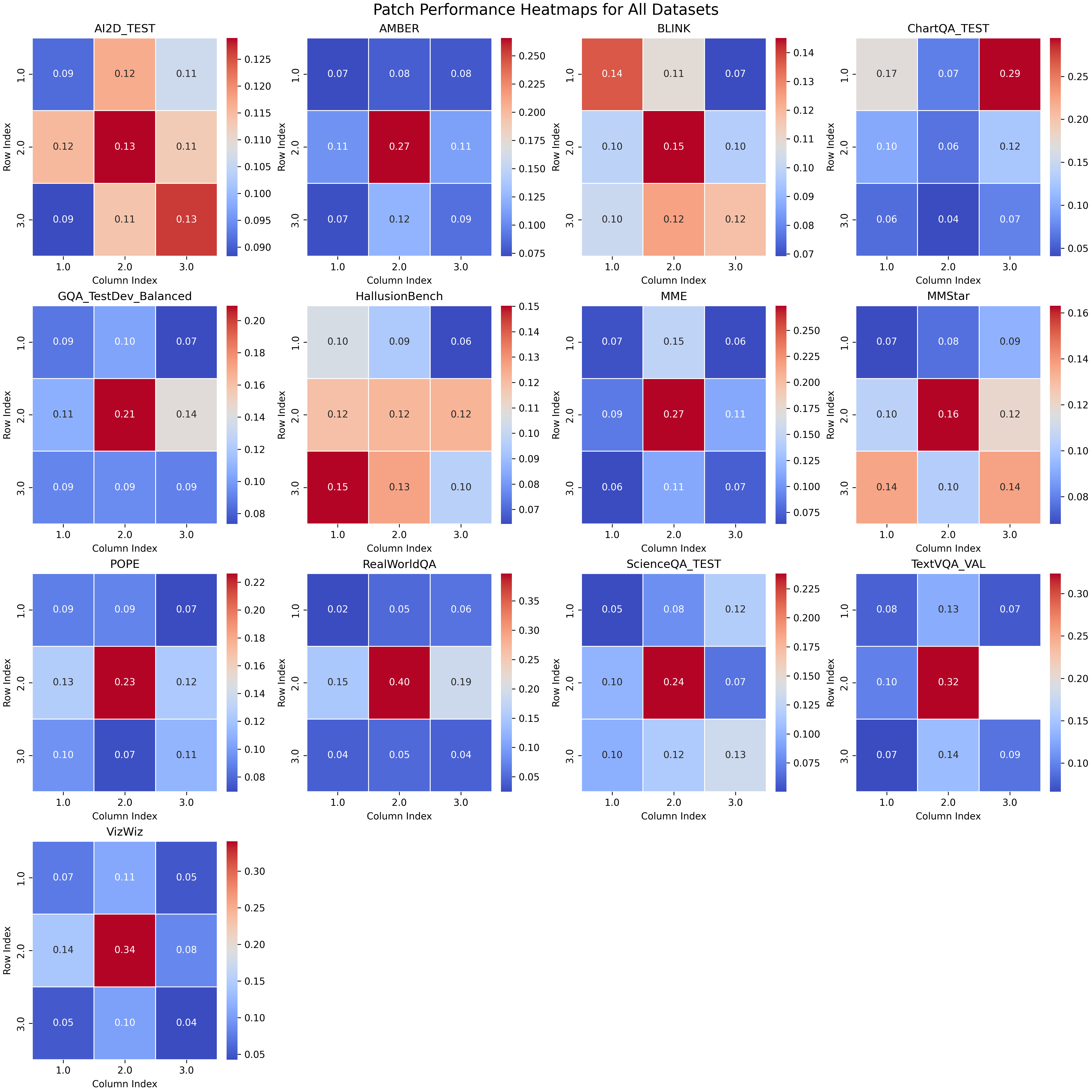}
    \caption{Patch-wise \(\text{MPP}_{n=3}\) performance heatmaps  across datasets. The results highlight that central and middle-row patches consistently dominating performance. Peripheral patches contribute significantly less, confirming that benchmarks favor spatial regions and localized reasoning, potentially leading to shortcut learning in multimodal models.}
    \label{fig:heatmap_patch_3}
\end{figure*}

\paragraph{Patch Performance Heatmaps (\( n=2 \) and \( n=3 \)).}
Figures~\ref{fig:heatmap_patch_2} and~\ref{fig:heatmap_patch_3} show heatmaps for patch-wise MPP contributions at both \( n=2 \) (2x2) and \( n=3 \) (3x3) granularities, for each benchmark. Color intensity indicates the relative importance of each patch position, averaged across all samples. Across most benchmarks, a pronounced central bias is evident at \( n=3 \), where the center patch (patch 5) accounts for a disproportionately large fraction of performance. In contrast, peripheral patches (e.g., corners and edges) are consistently underutilized, reinforcing the claim that current datasets often fail to require models to attend to the full spatial context.

\paragraph{Dataset-Specific and Granularity Insights.}
Certain benchmarks (e.g., ChartQA\_TEST, MMStar) display task-specific spatial dependencies, such as top-right or lower-left patches dominating due to consistent layout of legends or labels. Comparing \( n=2 \) with \( n=3 \), finer granularity reveals sharper biases that are masked at coarser levels, confirming that spatial artifacts are best diagnosed with higher patch resolution (balanced against computational feasibility).

\paragraph{Interpretation and Takeaways.}
These visualizations concretely demonstrate the prevalence of spatial shortcuts in widely used benchmarks and highlight how RCI and patch-based analysis enable dataset creators to detect and correct these biases. Practitioners should leverage such analyses during dataset development and choosing benchmarks to evaluate models, to ensure that benchmarks reflects the usecase requirements , thereby supporting the construction of more robust and generalizable multimodal models.
% require holistic, or localized, reasoning, thereby supporting the construction of more robust and generalizable multimodal models.

\section{Practical Implications for Industry and Deployment}
\label{sec:practitioner_appendix}

The Region Comprehension Index (RCI) offers a concrete, actionable tool for practitioners designing, auditing, and maintaining multimodal benchmarks and datasets for real-world applications. Integrating RCI into data and system workflows can provide the following practical benefits:

\begin{table*}[t]
\centering

\scalebox{1}{
\begin{tabular}{p{3.6cm} p{2.3cm} p{7.2cm}}
\toprule
\textbf{Application Domain} & \textbf{Recommended RCI} & \textbf{Reasoning Requirement / Rationale} \\
\midrule
Autonomous Driving, Remote Sensing, Medical Diagnostics & High ($\gg 0$) & Requires holistic/global understanding of the full scene; critical information is distributed across image regions. \\
Document Intelligence, Chart QA, Visual Analytics & Medium to High ($\gtrsim 0$) & Relies on integrating spatially dispersed or multi-element content; benefits from enforcing global context reasoning. \\
Facial Recognition, Fine-Grained Visual Inspection, Defect Detection & Low ($\lesssim 0$) & Performance is determined by local visual cues; global scene often irrelevant or distracting. \\
Multi-Purpose, General VQA, Open-World Tasks & Medium (near $0$) & Mixed reasoning needs; desirable to avoid strong local or global bias for broad applicability and generalization. \\
\bottomrule

\end{tabular}
}
\caption{Application scenarios and recommended RCI ranges. RCI helps practitioners select or design datasets best suited to the real-world reasoning demands of their use case.}
\label{tab:application_examples}
\end{table*}

\begin{enumerate}
    \item \textbf{Application-Driven Dataset Selection:} RCI enables practitioners to select or curate datasets according to the actual reasoning requirements of their target use cases. For example, high-RCI datasets are best suited for tasks requiring holistic scene understanding (e.g., autonomous driving, document analysis, complex visual QA), while low-RCI datasets are appropriate for local cue-driven applications (e.g., facial authentication, manufacturing defect detection).
    
    \item \textbf{Benchmark Auditing and Dataset Quality Control:} By systematically quantifying spatial reasoning dependencies, RCI helps identify central or localized biases that may undermine generalization. This allows dataset creators to remediate these issues—by diversifying spatial placement of task-relevant content or augmenting underrepresented regions—before system deployment.
    
    \item \textbf{Guided Dataset Construction and Improvement:} During iterative dataset development, RCI can be used as a continuous feedback signal: practitioners can set RCI targets for new benchmarks and use the metric to monitor and correct emergent biases as more data is collected or annotated.
    
    \item \textbf{Deployment Robustness and Continuous Monitoring:} For deployed systems, RCI can inform post-deployment auditing. If system failures or unexpected behaviors are observed in production, re-evaluating datasets with RCI can reveal whether unintentional local/global biases have crept in over time, prompting corrective data interventions.
    
    \item \textbf{Actionable Workflow for Practitioners:}
    \begin{itemize}
        \item \textbf{Step 1:} Apply RCI to candidate datasets or benchmarks to measure their dependency on local vs.\ global reasoning.
        \item \textbf{Step 2:} Compare RCI profiles to the reasoning needs of your application (see Table~\ref{tab:application_examples} for guidance).
        \item \textbf{Step 3:} If mismatches are found, use RCI as a feedback signal for data augmentation, sample selection, or benchmark redesign.
        \item \textbf{Step 4:} Re-evaluate after each major update to ensure continued alignment with application requirements.
    \end{itemize}
\end{enumerate}

RCI thus empowers both researchers and industrial teams to build, evaluate, and maintain datasets and benchmarks that more accurately reflect and anticipate real-world demands. This enables the development of robust, generalizable multimodal systems that perform reliably outside controlled lab settings.

% \begin{table}[t*]
% \centering
% \caption{Application scenarios and recommended RCI ranges. RCI helps practitioners select or design datasets best suited to the real-world reasoning demands of their use case.}
% \label{tab:application_examples}
% \begin{tabular}{p{3.5cm} p{2.2cm} p{6cm}}
% \toprule
% \textbf{Application Domain} & \textbf{Recommended RCI} & \textbf{Reasoning Requirement / Rationale} \\
% \midrule
% Autonomous Driving, Remote Sensing, Medical Diagnostics & High ($\gg 0$) & Requires holistic/global understanding of the full scene; critical information is distributed across image regions. \\
% \addlinespace
% Document Intelligence, Chart QA, Visual Analytics & Medium to High ($\gtrsim 0$) & Relies on integrating spatially dispersed or multi-element content; benefits from enforcing global context reasoning. \\
% \addlinespace
% Facial Recognition, Fine-Grained Visual Inspection, Defect Detection & Low ($\lesssim 0$) & Performance is determined by local visual cues; global scene often irrelevant or distracting. \\
% \addlinespace
% Multi-Purpose, General VQA, Open-World Tasks & Medium (near $0$) & Mixed reasoning needs; desirable to avoid strong local or global bias for broad applicability and generalization. \\
% \bottomrule
% \end{tabular}
% \end{table*}

\subsection{A New Paradigm for Benchmark Evaluation}
\label{sec:impact}
The adoption of \( \text{RCI} \) introduces a new paradigm for quantifying multimodal dataset biases. Unlike traditional dataset metrics, which focus on linguistic alignment or classification accuracy, \( \text{RCI} \) explicitly measures spatial reasoning dependencies, providing a more interpretable framework for evaluating dataset structure.

\begin{table*}[!ht]
\centering
\renewcommand{\arraystretch}{1.2} % Adjust row height
\setlength\dashlinedash{0.5pt}    % Dash length
\setlength\dashlinegap{1pt}       % Gap between dashes
\setlength\arrayrulewidth{0.8pt}  % Thickness of outer border

\vspace{0.5em}

\scalebox{0.9}{
\begin{tabular}{|>{\centering\arraybackslash}m{3.8cm}
                |>{\centering\arraybackslash}m{2cm}
                |>{\centering\arraybackslash}m{2cm}
                |>{\centering\arraybackslash}m{2cm}
                |>{\centering\arraybackslash}m{3.8cm}
                |>{\centering\arraybackslash}m{2cm}|} 
\hline
\textbf{Question} & \textbf{Original} & \textbf{2×2 Split} & \textbf{3×3 Split} & \textbf{Answerable with} \\
\hline

Which direction is the arrow pointing? &
\includegraphics[height=2cm]{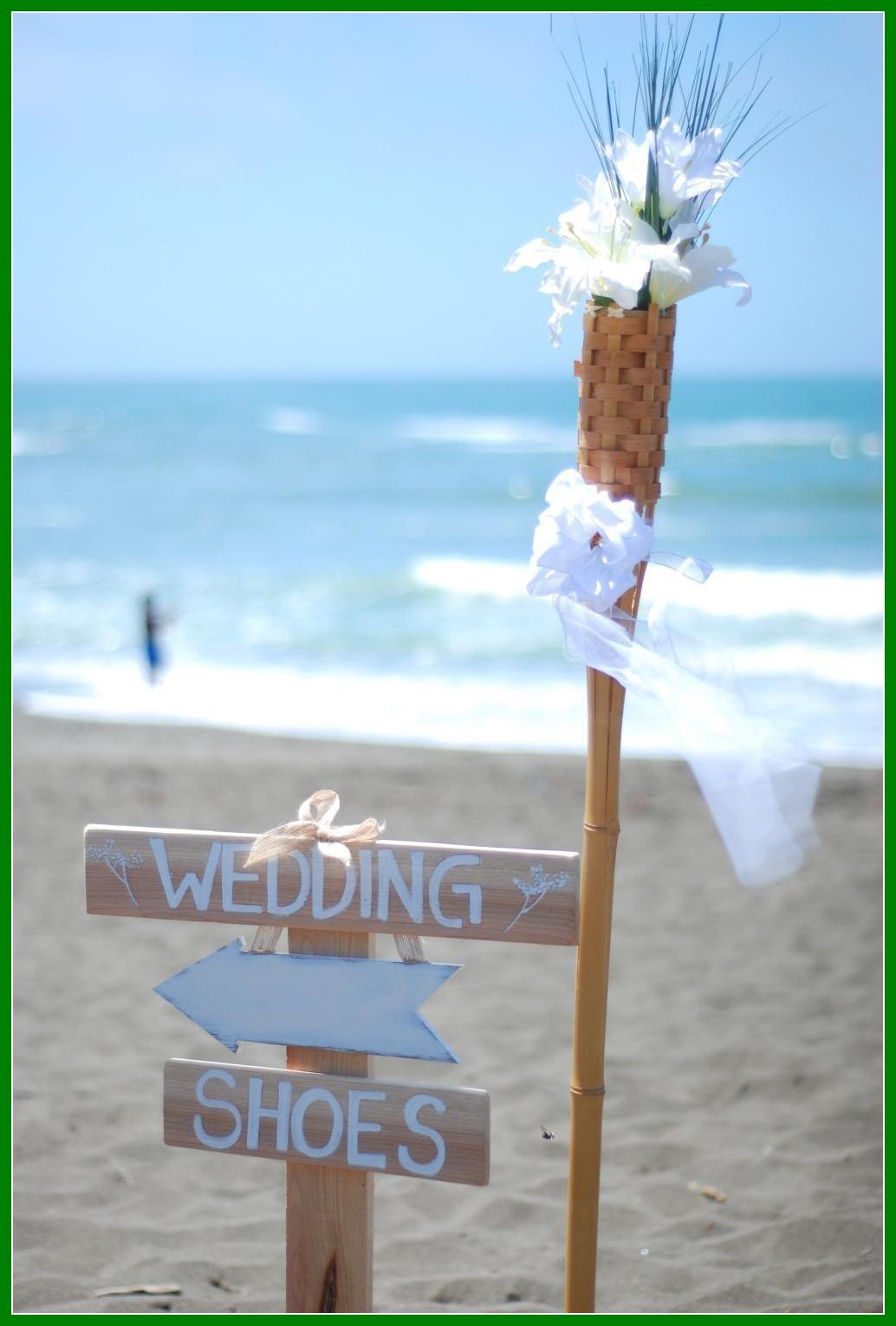} &
\includegraphics[height=2cm]{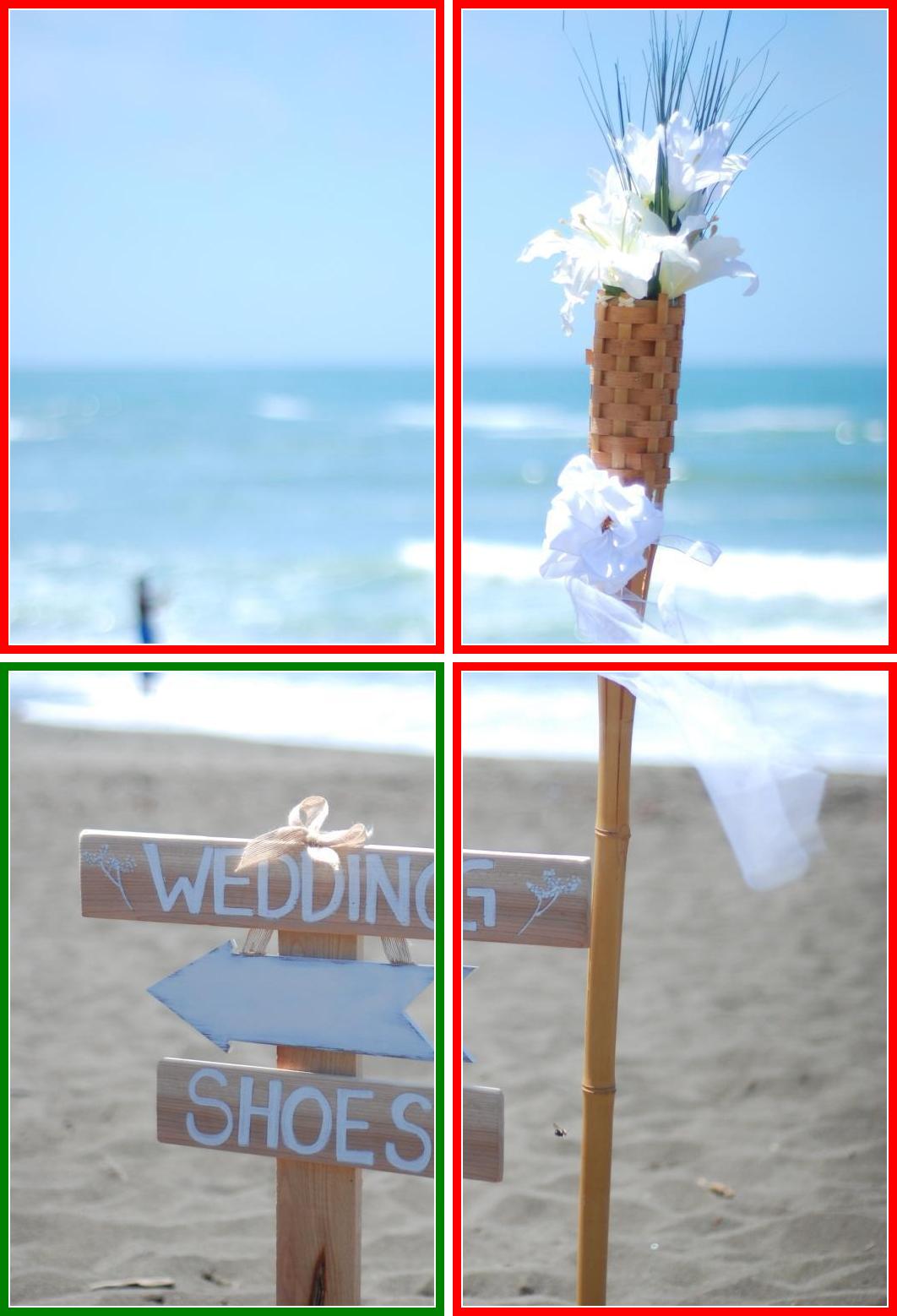} &
\includegraphics[height=2cm]{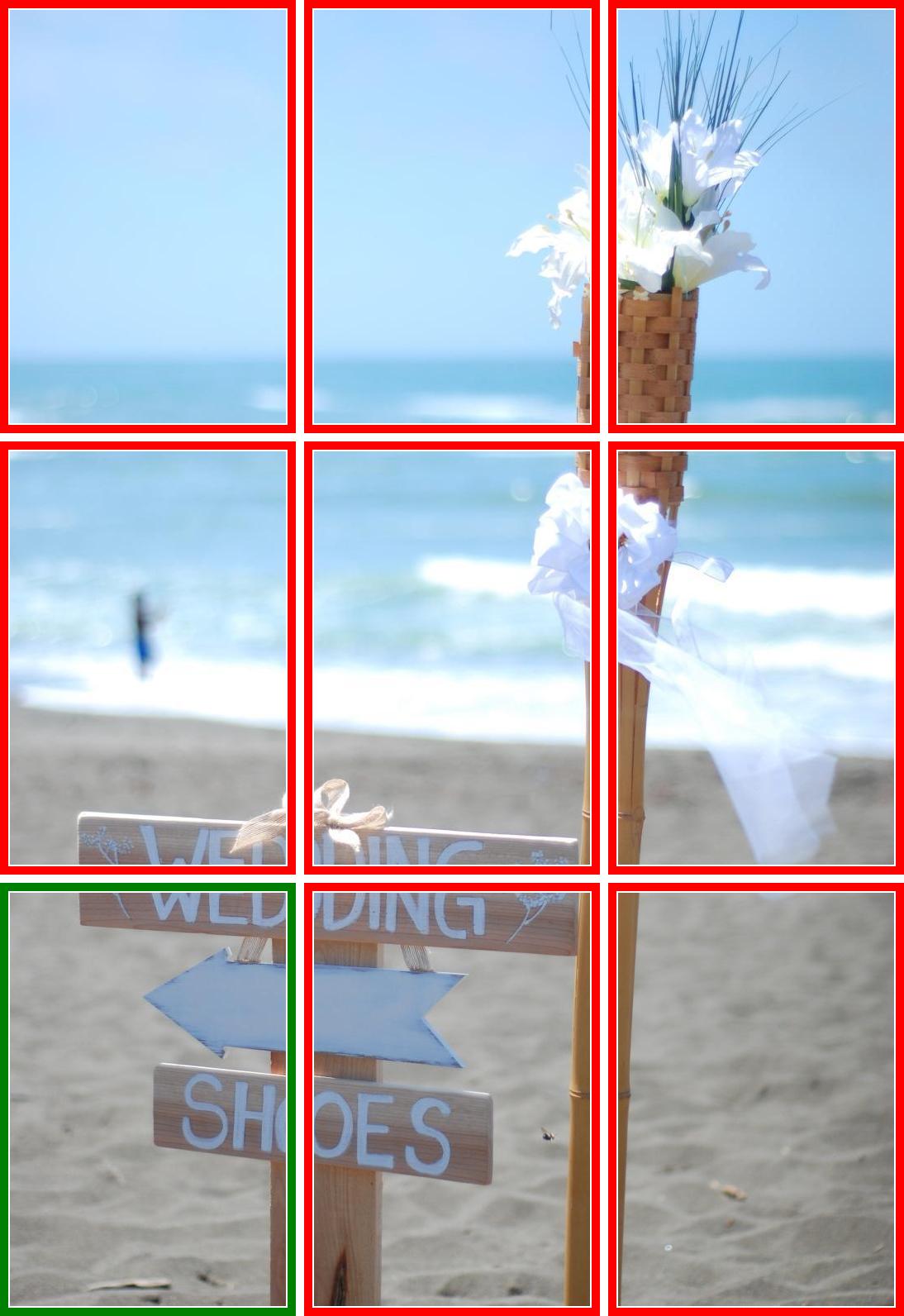} &
Full image, 2×2 split (row 2, column 1), 3×3 split (row 3, column 1) \\
\hline

How many stamps are in this photo? &
\includegraphics[height=2cm]{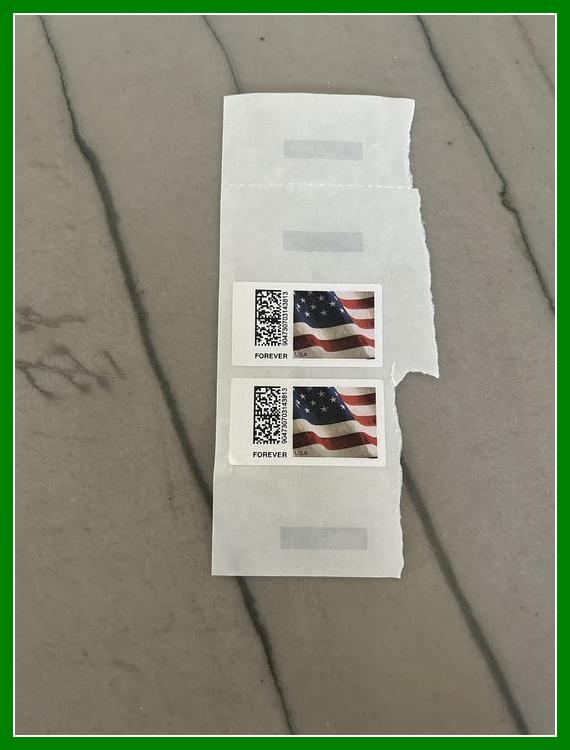} &
\includegraphics[height=2cm]{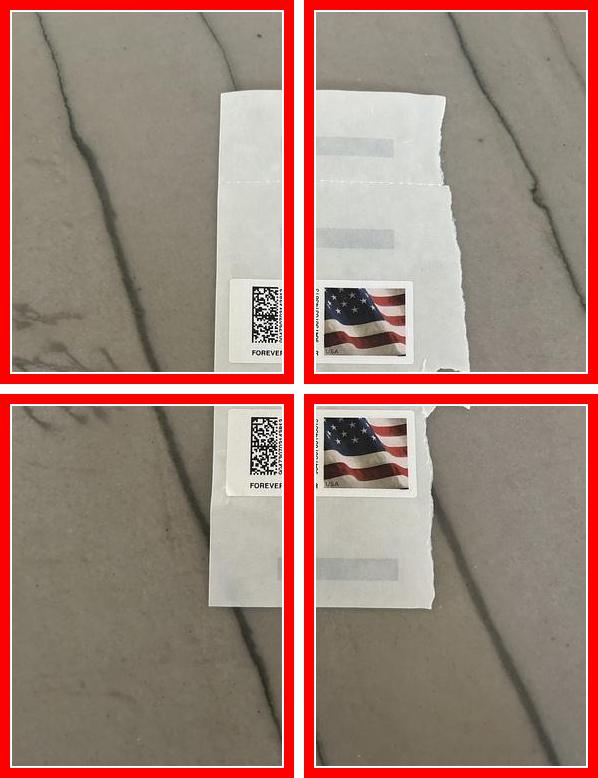} &
\includegraphics[height=2cm]{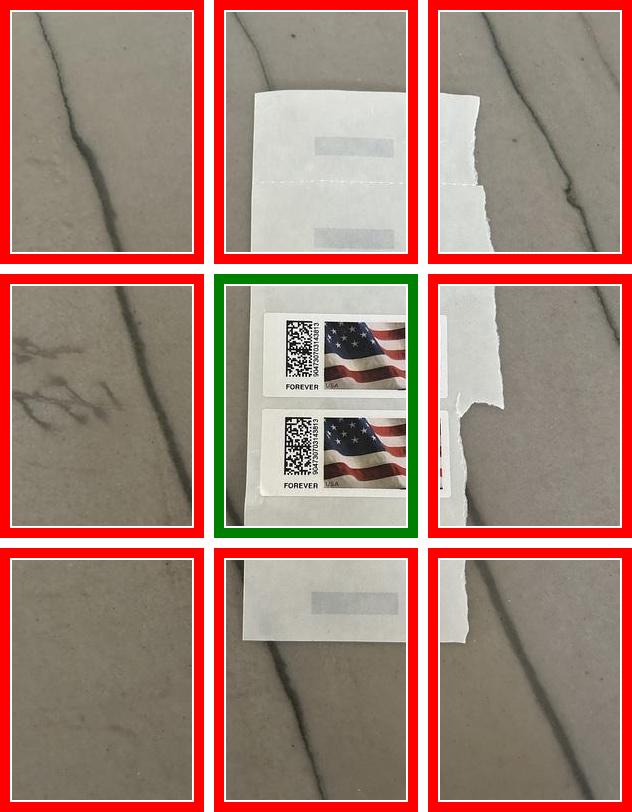} &
Full image, 3×3 split (row 2, column 2) \\
\hline

Where is the toy relative to the dog? &
\includegraphics[height=2cm]{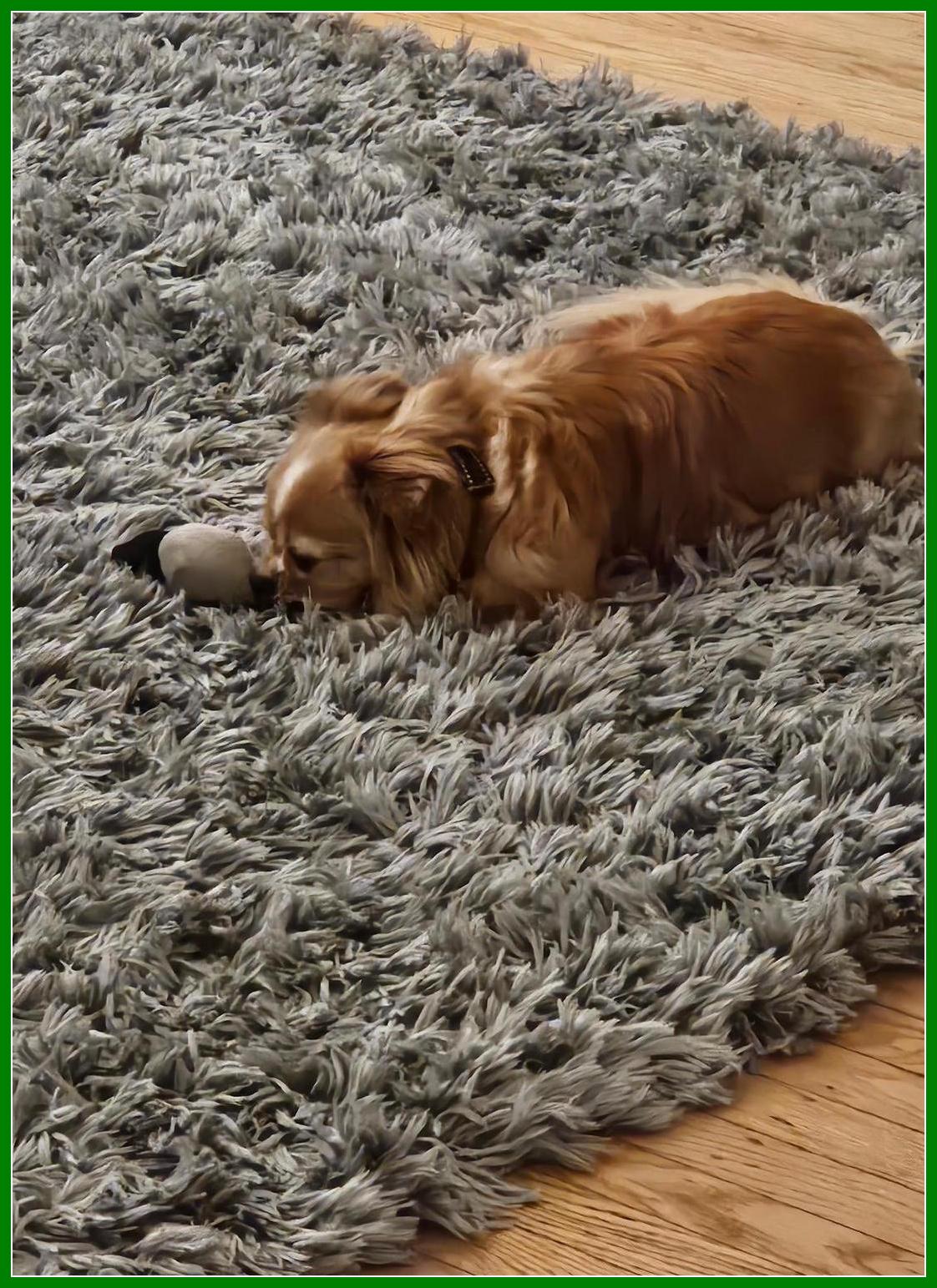} &
\includegraphics[height=2cm]{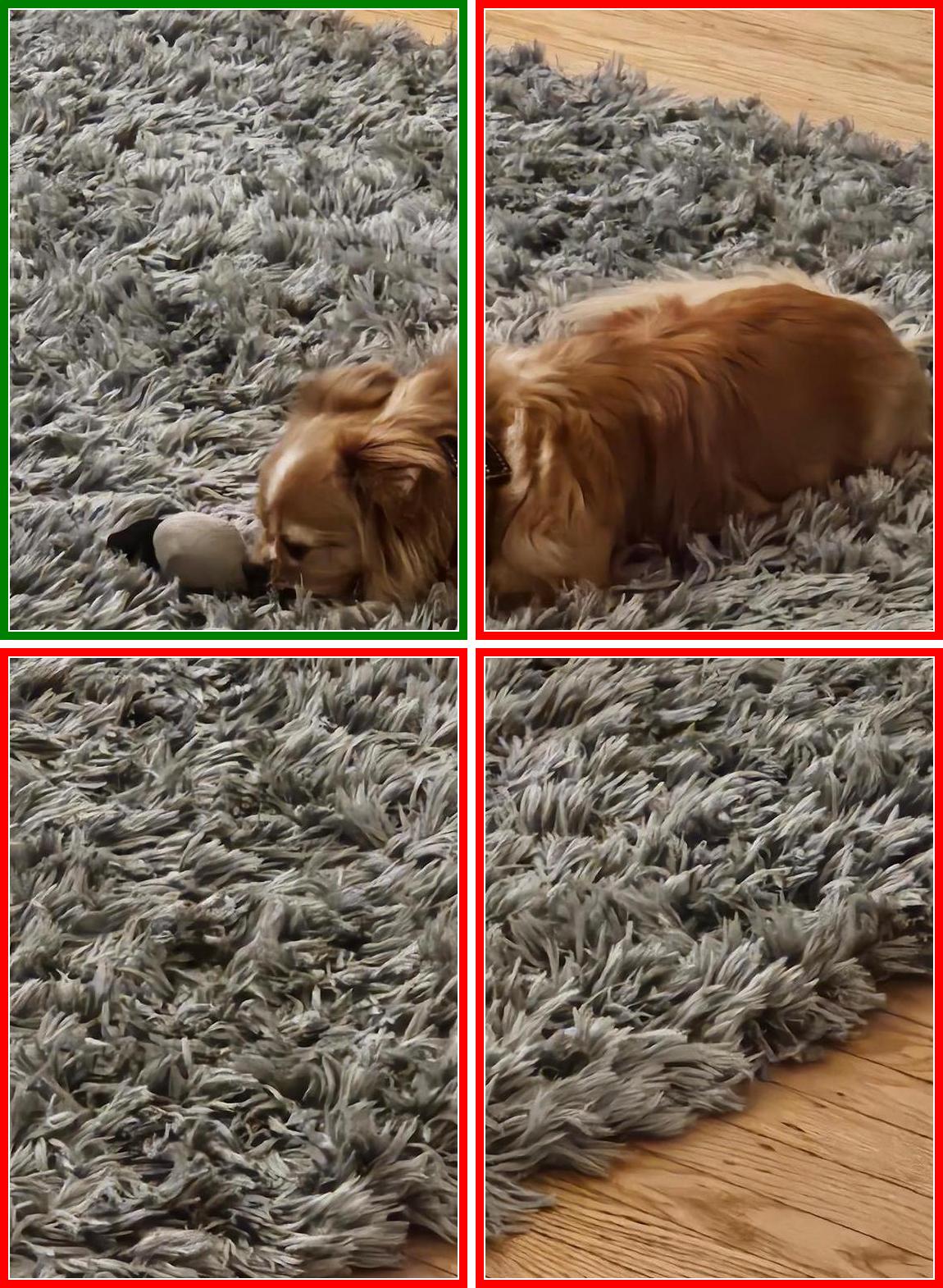} &
\includegraphics[height=2cm]{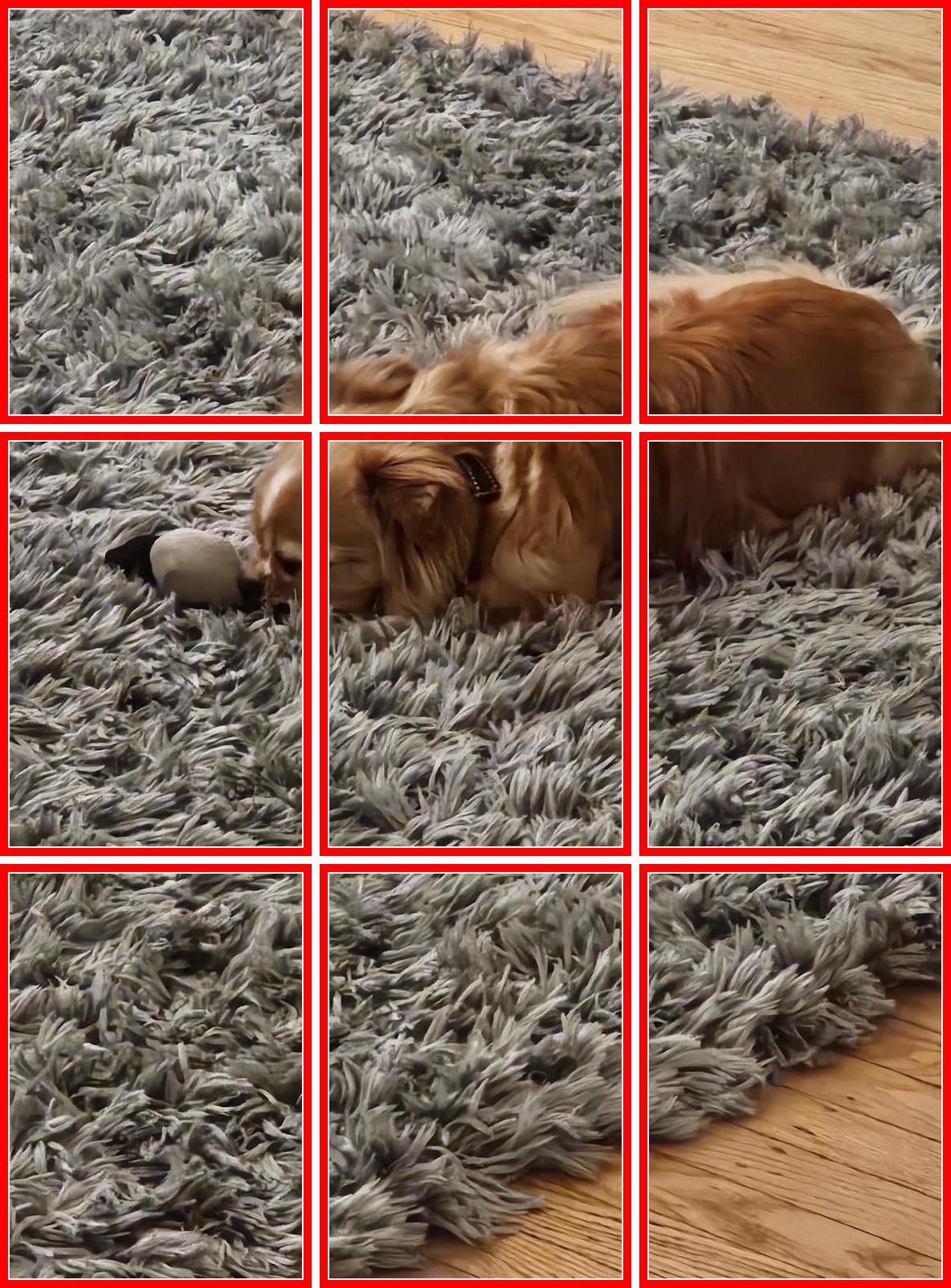} &
Full image, 2×2 split (row 1, column 2) \\
\hline

Which cat is smaller? &
\includegraphics[height=2cm]{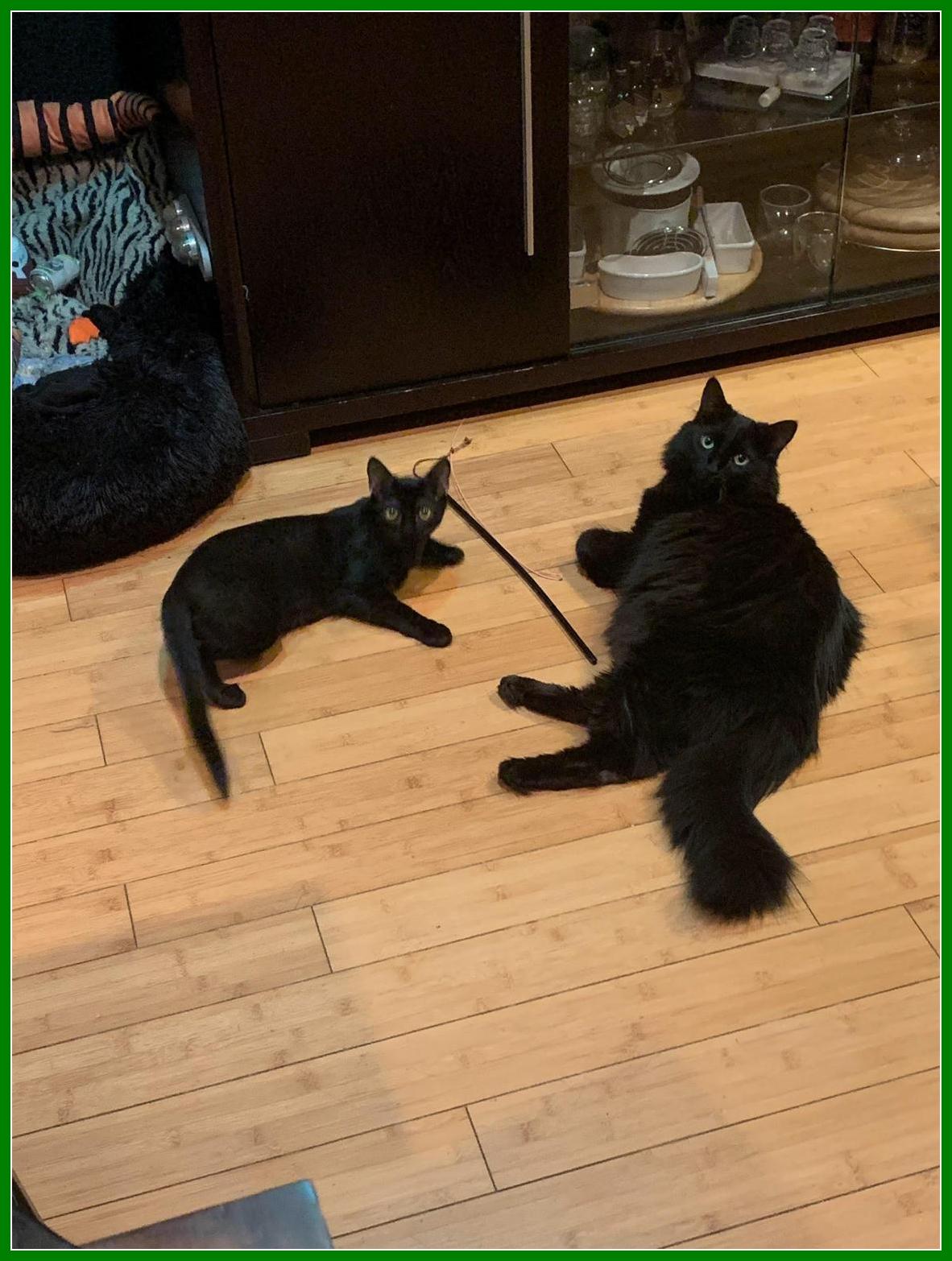} &
\includegraphics[height=2cm]{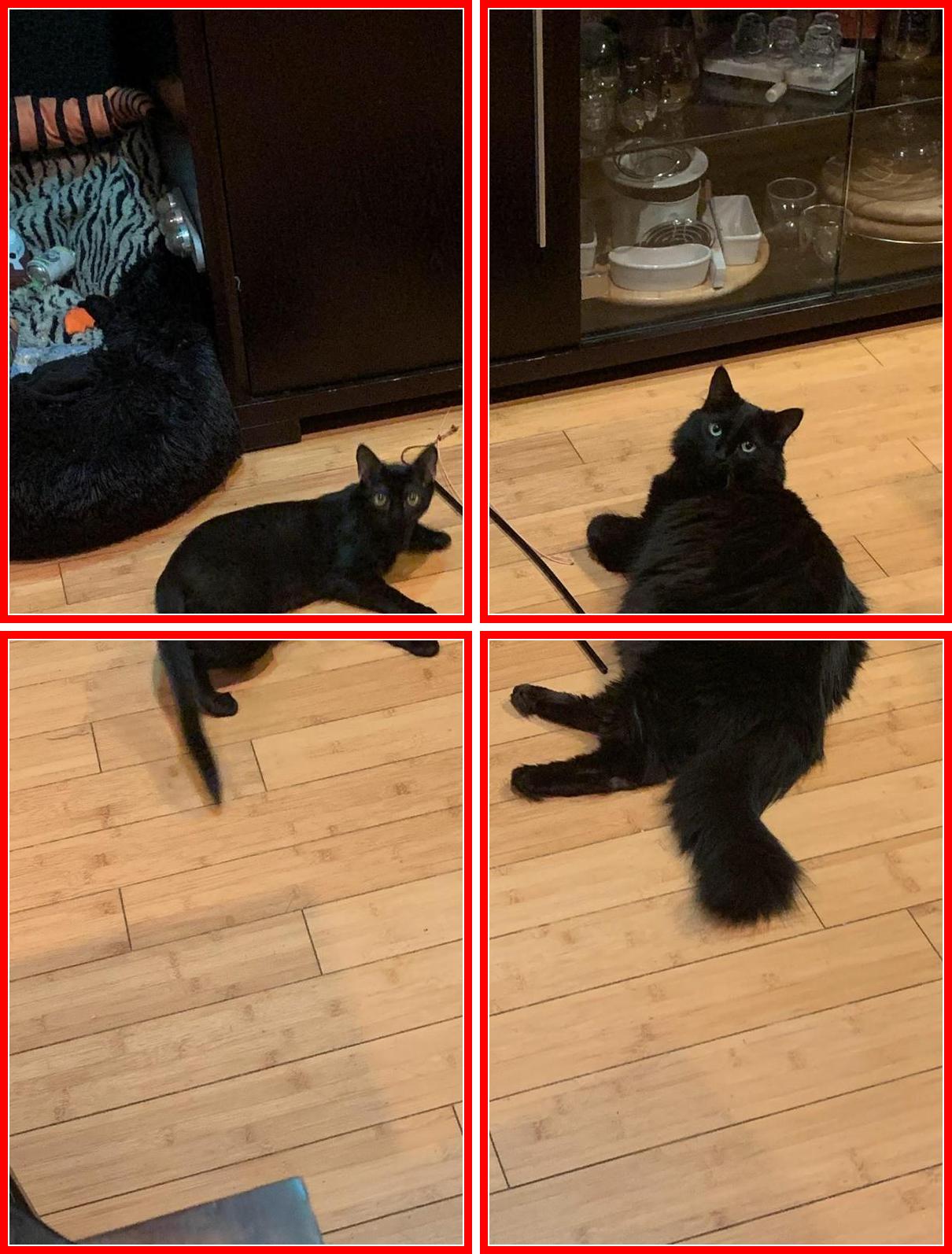} &
\includegraphics[height=2cm]{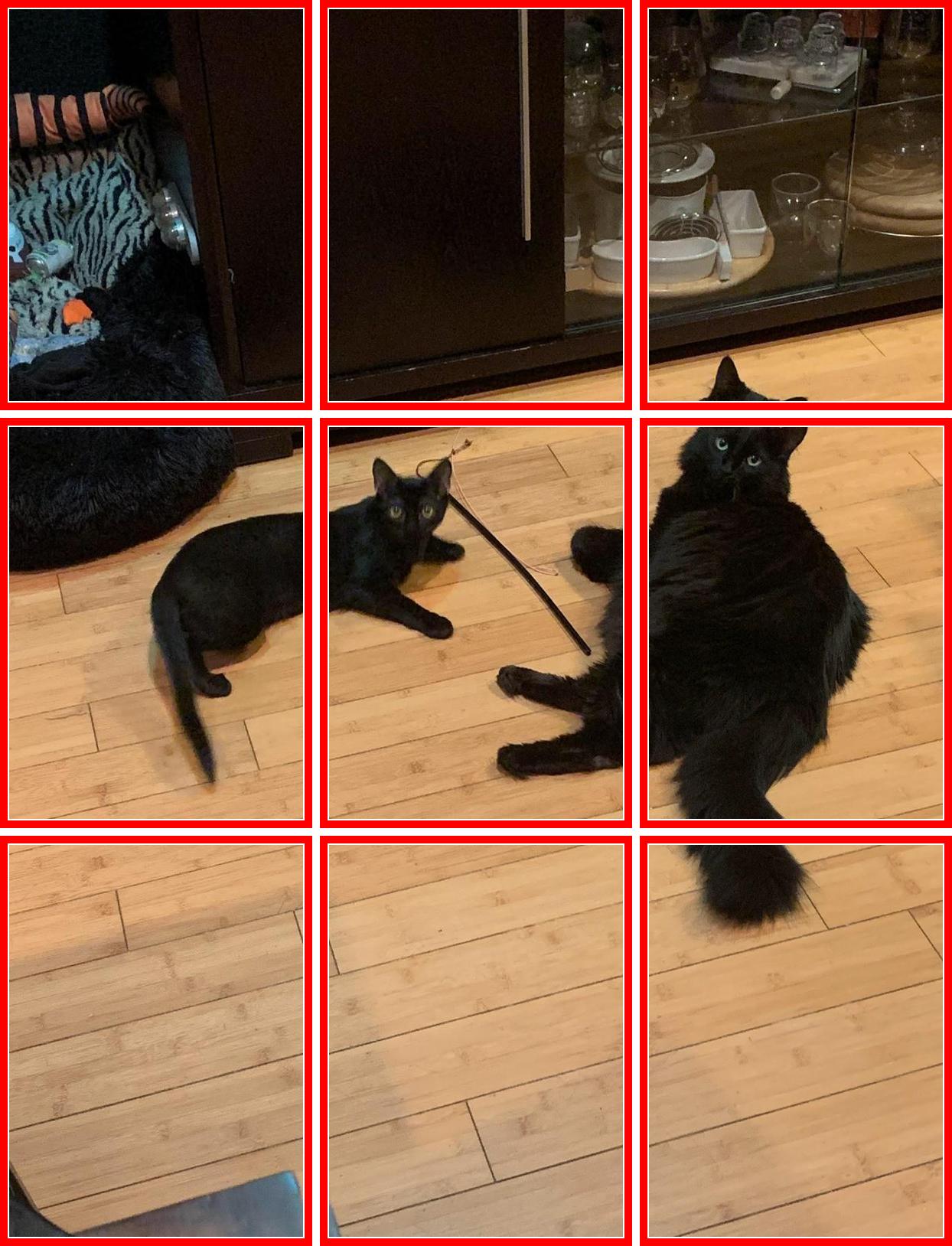} &
Full image \\
\hline
\end{tabular}
}
\caption{
Illustrative examples from the RealWorldQA dataset demonstrating how context granularity (full image vs. 2×2 and 3×3 splits) affects answerability. For each question, the answerable regions are highlighted, showing which image segments provide sufficient context for accurate answering. \textcolor{green}{Green} boundaries denote sufficient context; \textcolor{red}{Red} boundaries indicate insufficiency.
}
\label{tab:realworldqa_rci_examples}
\vspace{-1em}
\end{table*}

This paradigm shift has three major implications:
\begin{itemize}
    \item Researchers can use \( \text{RCI} \) to systematically quantify reasoning requirements, revealing dataset biases that traditional metrics overlook.
    \item Benchmark designers can apply \( \text{RCI} \) to ensure datasets enforce distributed visual reasoning rather than localized shortcut learning.
    \item Practitioners can leverage \( \text{RCI} \) to identify which datasets to train, benchmarks and models are best suited for real-world multimodal applications.
\end{itemize}

By setting a higher standard for training and benchmarking datasets, this work aims to drive progress toward more interpretable, robust, and generalizable vision-language models.

\subsection{Real-World Relevance}
\label{sec:real_world}
The biases revealed by \(\text{RCI}\) have significant implications for real-world applications of vision-language models:
\begin{itemize}
    \item \textbf{Complex Environments.} Applications such as autonomous driving, medical imaging, and robotics require global reasoning across distributed visual information. Models trained on biased datasets may struggle to generalize to these contexts.
    \item \textbf{Robustness and Interpretability.} Models that rely on localized cues are less robust to unseen scenarios and harder to interpret. Incorporating benchmarks with higher \(\text{RCI}\)'s can lead to more reliable and interpretable systems.
    \item \textbf{Region Comprehension in Real-Time Systems.} Tasks such as disaster response and surveillance demand real-time processing of spatially distributed information. Evaluating and improving region comprehension is critical for such high-stakes scenarios.
\end{itemize}

\subsection{Illustrative Examples}

\label{sec:example_RCI}
To qualitatively demonstrate how RCI reflects visual reasoning dependencies in vision-language tasks, Table~\ref{tab:realworldqa_rci_examples} presents representative examples from the RealWorldQA benchmark. Each row shows a question along with the original image, its 2×2 and 3×3 patch splits, and visual context needed to answer the question. These examples highlight varying degrees of reliance on local versus global visual information.

These examples demonstrate how RCI serves as a fine-grained diagnostic tool to assess the extent of visual context needed for accurate question answering, helping uncover local vs. global reasoning biases in multimodal datasets

\section{Limitations and Future Work}
\label{sec:limitations}

While \( \text{RCI} \) provides a novel framework for evaluating dataset biases and spatial reasoning requirements, several limitations remain. Addressing these challenges will be essential for expanding its applicability and ensuring comprehensive benchmark evaluation.

\subsection{Current Limitations}
\paragraph{Model Dependency.} 
Since \( \text{RCI} \) is computed using model-based evaluations, its results may be influenced by model architecture and training biases. Future work should explore the use of model ensembles or develop a standardized reference model set to ensure consistency across evaluations.

\paragraph{Single-Image Focus.} 
Our evaluation is primarily suited for single-image benchmarks, limiting its applicability to datasets requiring multi-image context or video-based reasoning. Extending \( \text{RCI} \) to sequential data would enable an understanding of temporal biases and multi-frame dependencies.

\paragraph{Task-Specific Sensitivity.} 
Certain tasks, such as single-object detection, inherently rely on localized information rather than distributed reasoning. Future adaptations of \( \text{RCI} \) should account for task-specific dependencies, ensuring fair evaluations across different types of multimodal tasks.

% \paragraph{Lack of Model-Independent Evaluation.}  
% \( \text{RCI} \) currently relies on pretrained models to assess dataset biases. Developing additional model-agnostic evaluation techniques, such as feature-space clustering or entropy-based spatial analysis, could provide complementary insights independent of model architectures.

\subsection{Future Research Directions}
\paragraph{Expanding to Multi-Image and Video Understanding.}  
To evaluate spatial biases in multi-frame reasoning, we propose extending \( \text{RCI} \) to video datasets and multi-image benchmarks. This would allow for analyzing how dataset biases evolve over time, particularly in tasks requiring temporal consistency.

\paragraph{\( \text{RCI} \) for Dataset Optimization.}
Beyond evaluating existing benchmarks, \( \text{RCI} \) can be leveraged for training dataset construction and filtering. By computing \( \text{RCI} \) at the dataset level, we can:
\begin{itemize}
    \item Filter out samples where models can succeed using only localized information, ensuring a dataset composition that requires global contextual reasoning.
    \item Design adversarial datasets that distribute task-relevant information across multiple regions, preventing models from relying on shortcuts.
    \item Implement automated dataset rebalancing, where training data selection is adjusted dynamically based on \( \text{RCI} \).
\end{itemize}

% \paragraph{Automated Dataset Rebalancing.}  
% Our findings confirm that certain datasets consistently reinforce central bias. Future work should explore automated dataset rebalancing algorithms that dynamically redistribute key information across different image regions, ensuring fairer multimodal evaluation.

\paragraph{Adversarial Benchmarking.}  
To prevent models from exploiting dataset biases, we propose designing adversarial dataset variants. These datasets would include:
\begin{itemize}
    \item Randomized spatial layouts to test whether models generalize beyond static object placements.
    \item Occlusion-based modifications to analyze how models adapt when key visual cues are hidden.
    \item Cross-domain shifts to evaluate model robustness in unseen distributions.
\end{itemize}

% \paragraph{Cross-Modality Spatial Reasoning.}  
% While \( \text{RCI} \) evaluates image-based benchmarks, its adaptation to text-image datasets could enable a deeper understanding of how spatial biases influence multimodal reasoning. Future research should explore how textual descriptions interact with spatial image representations, particularly in vision-language grounding tasks.

By addressing these challenges, \( \text{RCI} \) can evolve into a comprehensive framework across different modalities not only for dataset evaluation but also for dataset and model optimization. Future research can explore integrating \( \text{RCI} \) directly into the training pipeline to improve dataset fairness, balance spatial reasoning, and enhance multimodal model generalization, ensuring more robust and interpretable benchmarking across multimodal AI research.

\end{document}